\documentclass[reqno]{amsart}
\usepackage{amsmath}
\usepackage{amsopn}
\usepackage{amssymb}
\usepackage{amsfonts}
\usepackage{graphicx}
\usepackage{boldline}
\usepackage{graphicx,booktabs}
\usepackage[table]{xcolor}
\usepackage{subfigure}
\usepackage{fancyhdr}
\usepackage[section]{placeins}
\usepackage [autostyle, english = american]{csquotes}
\usepackage{multicol}
\usepackage{booktabs,dcolumn,caption}
\usepackage{url}
\usepackage[colorlinks,allcolors=blue]{hyperref} 
\numberwithin{table}{section}
\usepackage{subfigure}
\usepackage{makecell}
\usepackage{lscape} 
\usepackage{multirow}
\usepackage{longtable}

\newcommand{\rr}{\raggedright}
\newcommand{\tn}{\tabularnewline}

\newcommand{\rc}{\centering}
\newcolumntype{R}[1]{>{\raggedleft\let\newline\\\arraybackslash\hspace{0pt}}m{#1}}
\newcolumntype{L}[1]{>{\raggedright\let\newline\\\arraybackslash\hspace{0pt}}m{#1}}
\newcolumntype{C}[1]{>{\centering\let\newline\\\arraybackslash\hspace{0pt}}m{#1}}

\usepackage[foot]{amsaddr}

\definecolor{maroon}{cmyk}{0,0.87,0.68,0.32}

\begin{document}

\title{\textbf{LS\MakeLowercase{c}DC -- New Large Scientific Dictionary }}

\author{N. S\"UZEN$^{1}$}
\author{E. M. MIRKES$^{1,2}$}
\author{A. N. GORBAN$^{1,2}$}

\address{$^{1}$University of Leicester, Leicester LE1 7RH, UK}
\address{$^{2}$Lobachevsky University, Nizhni Novgorod, Russia}

\email{ns433@leicester.ac.uk (N. Suzen)}
\email{em322@leicester.ac.uk (E.M. Mirkes)}
\email{ag153@leicester.ac.uk (A.N. Gorban)}

\maketitle

\begin{abstract}

In this paper, we present a scientific corpus of abstracts of academic papers in English -- Leicester Scientific Corpus (LSC). The LSC contains 1,673,824 abstracts of research articles and proceeding papers indexed by Web of Science (WoS) in which publication year is 2014. Each abstract is assigned to at least one of 252 subject categories. Paper metadata include these categories and the number of citations. We then develop scientific dictionaries named Leicester Scientific Dictionary (LScD) and Leicester Scientific Dictionary-Core (LScDC), where words are extracted from the LSC. The LScD is a list of 974,238 unique words (lemmas). The LScDC is a core list (sub-list) of the LScD with 104,223 lemmas. It was created by removing LScD words appearing in not greater than 10 texts in the LSC. LScD and LScDC are available online. Both the corpus and dictionaries are developed to be later used for quantification of meaning in academic texts.  
    
Finally, the core list LScDC was analysed by comparing its words and word frequencies with a classic academic word list `New Academic Word List (NAWL)' containing 963 word families, which is also sampled from an academic corpus. The major sources of the corpus where NAWL is extracted are Cambridge English Corpus (CEC), oral sources and textbooks. We investigate whether two dictionaries are similar in terms of common words and ranking of words. Our comparison leads us to main conclusion: most of words of NAWL (99.6\%) are present in the LScDC but two lists differ in word ranking. This difference is measured.   

\vspace{5mm}


\end{abstract}

\tableofcontents

\section{Introduction}
\subsection{Quantification of Meaning in Academic Texts\nopunct}  \hspace*{\fill} \\ 

The interest of adaptation the modern technologies to text mining is growing fast, along with the awareness of the importance of textual data in almost all industries. The increase in the number of users of the internet and social media platforms makes a huge amount of textual data available that play a crucial role on research and marketing strategies. 

The storage of almost all types of data in electronic platforms and the spread of the social networking sites for scientists open up opportunities for researchers to share scientific researches and to access a wide range of publication repositories freely and effectively. According to \cite{rg1,rg2}, the largest academic social network ResearchGate has 15+ million researchers registered, with a huge number of publications in multidisciplinary journals. The problem of searching for relevant papers out of an enormous number of publications and extraction the information from texts gradually becomes crucial. Therefore, automated procedure of text processing and extracting the meaning of a text have become important issues.

In Natural Language Processing (NLP) and computational linguistics, formal identifying `which meaning a text includes’ is still an open problem. Although in standard text mining applications one may relate the problem to ‘topic identification’ or ‘determining the class of text’, we consider this problem more widely. Our goal is not only determining ‘which class (classes) a text belongs to’ or ‘the topic (content) of a text’, but also numerical representing the meaning of the text. 
 
This task is different from quantification in classification applications. \textit{Quantification (or text quantification)} in classification is defined as the activity of estimating the prevalence (relative frequency) of each class in a set of unlabelled items \cite{sebastiani}. The aim here is, given a training set with class labels, to induce a quantifier that takes unlabelled test set, and then accurately estimate the number of cases in each class. Rather than estimating the prevalence of classes (or different meanings) in the corpus or determining the topic of the paper, we intend to represent the meaning in a research text numerically — so-called \textit{quantification of meaning in research texts}. Our assumption is that words and texts have meaning priors and meaning of a text can be extracted, at least, partially, from the information of words for subject categories distributed over the corpus. Given such information, these priors can be exploited firstly for each word and then each research text which is a collection of words. In other words, meaning of a research text is generated when different bits of information are associated with subject categories \cite{welbers}.
 
This approach follows the classical psycholinguistic ideas of measurement of meaning \cite{osgood} but instead of psychologically important sentiments the research categories are used. Each word is represented as a vector of information scores for various categories, and the meaning of the whole text is formalised as a cloud of these vectors for words from the text. For larger texts this approach can take into account correlations between words (co-occurrence of words and combinations of words). 
 
Quantities of meanings of texts can be later used in a number of applications involving categorisation of texts to pre-existing categories, creation of `natural' categories or more precisely, clustering of similar texts in such a way that texts in the same group have same/similar meanings. The solution to the problem of ‘quantification of meanings in text’ also impacts on other issues such as prediction of success of a scientific paper. 

Let us consider, for instance, grouping texts based on their contents. Bringing related research texts together gives the community a convenient and easily accessible location where a deep digging becomes possible inside. This provides users many benefits such as learning the hottest topics, the most significant researches and the latest developments in a specific field. Such automated mechanisms have also benefit for editors to help them in associating researches, for instance in the step of evaluating a new submission to determine whether it fits to the journal and standards in the field in terms of content, and more importantly to initiate the peer review process by selecting experts in the field.

In practice, searching and preliminary express understanding of a paper's content is generally done by reading title and abstract rather than reading full-text of the paper. Therefore, it is reasonable to search for relevant papers by searching of relevant abstracts. Natural questions needed to be answered here are: how to automatically process the abstracts, extract the meaning from such relatively short texts and represent the meaning in a text numerically to make them usable for text mining algorithms. These questions are some of our focuses to be answered through our research. 

This work is the first stage in the project outlined. In this paper, we consider the creation of an academic corpus and scientific dictionaries where words are extracted from the corpus. All steps in the creation process will be presented in details. The corpus and dictionaries will be used for quantification of meaning in academic texts in later stages of our research. 

\subsection{Building a Scientific Corpus: Leicester Scientific Corpus (LSC)\nopunct}  \hspace*{\fill} \\ 

One of the key issues in the analysis of meaning of texts is to use a corpus that is built in accordance with the scope of the study. Quantification of meaning of research texts, extracting information of words for subject categories and later prediction of success of the paper using quantity of texts naturally require well-organised, up-to-date and annotated corpus by subject categories and the information of citations along with the abstracts. For this purpose, we developed a \textit{scientific corpus} where texts are abstracts of research papers or proceeding papers, followed by creation of scientific dictionaries where words are extracted from the corpus. In this paper, we focus on building of a scientific corpus and dictionary to be used in future work on the quantification of the meaning of research texts. All steps of creation the corpus and the dictionary are presented in the later sections of the paper. 

The Leicester Scientific Corpus (LSC) is a collection of 1,673,824 English written abstracts of research articles and proceeding papers indexed by Web of Science (WoS) \cite{wos}, selected so as to represent the largest variety of abstracts of scientific works published in 2014 \cite{lsc}. Texts within the corpus are distributed across 252 subject categories -- with over 298 million words including stop words. No consideration is given to the selection of categories, we extracted all texts regardless of how many texts are included in an individual category. Each document in the corpus includes the text of abstract and the following metadata: title, list of subject categories, list of research areas, and times cited \cite{woscat,wosra,wostc}. Documents also have the list of authors with the exception of 119 documents, we did not exclude these documents. We collected documents in July 2018; therefore, the number of citations is from the publication date to July 2018.  

Given the LSC, we also intend to create scientific dictionary where words are extracted from the texts of abstracts in LSC to be used on measuring the information of words for subject categories in the process of quantification of meaning of texts. To better represent scientific fields, the variety of disciplines and corpus size are very important criteria in dictionary creation. The more disciplines where texts are collected from, the bigger and comprehensive dictionary can be created. Similarly, the more articles are collected, the more representative set of words for specific fields can be gathered. As we did not exclude any category from the corpus of 1,673,824 texts distibuted over 252 categories, we expect a reasonable representativeness of words. In addition, the dynamic nature of languages and changes of words with new discoveries -- due to fast changes in science and technology -- lead to a need for an up-to-date scientific dictionary. Thus, we created two scientific dictionaries based on a large, multidisciplinary corpus of academic English: Leicester Scientific Dictionary (LScD) and Leicester Scientific Dictionary-Core (LScDC) \cite{lscd,lscdc}.  

\subsection{Building Scientific Dictionaries: Leicester Scientific Dictionary (LScD) and Leicester Scientific Dictionary-Core (LScDC)\nopunct}  \hspace*{\fill} \\ 

LScD is a list of words extracted from the texts of abstracts in the LSC. The words in the LScD is sorted by the number of documents containing the word in descending order. There are 974,238 unique words (lemmas) in the dictionary. All words in the LScD are in stemmed form and stop words are excluded. The dictionary also contains the number of documents containing each word and the number of appearance of the word in entire corpus. All steps to process the LSC, build LScD and the basic statistics with characteristics of words in the LScD are presented in the later sections of this paper.

LScDC is a core list (sub-list) of the LScD. The dictionary contains of 104,223 unique words (lemmas). The following decision is taken in the creation of the LScDC: words (in LScD) that appearing in not greater than 10 documents ($\leq 10$) are removed under the assumption that too rare words are not informative for text categorisation and gives almost zero scores in the calculation of information score as they appear in less than 0.01\% of documents. 60\% of words in LScD appear in only one document. Our casual observation of words indicates that many of such words are non-words or not in an appropriate format to use (e.g. misspelling); therefore, they are likely to be non-informative signals (noise) for algorithms. More information and examples can be found in Section \ref{lscdsec} and Section \ref{lscdcsec}. Removal of such words results in reducing the number of words in applications of text mining algorithms. When the threshold 10 is decided, we consider a cut which is not too small or high to be able to keep a reasonable number of words for analysis, but we paid attention to have a noticeable impact on size of dictionary and results. We did not remove any frequent words in this stage as stop words are already removed in pre-processing steps. The core dictionary is also ordered by the number of documents containing the words and includes the information of the number of appearance of the word in entire corpus.  

\subsection{A Comparison of the LScDC and the New Academic Word List (NAWL) \nopunct}  \hspace*{\fill} \\ 

This study also compares the LScDC and the New Academic Word List (NAWL) \cite{browne2}. The procedure used to compare involves looking at a classic academic list of words NAWL and investigating whether two lists contain the same words, the number of common words, possible reason of mismatches and whether ratings of matched words are actually the same/similar in two dictionaries. Overall, we intend to see whether there is similarity between two lists.

The reason why we consider the NAWL for comparison lies in two facts. The major reason lies in the way the sampling of the vocabulary, which is similar to ours in terms of being from a general and academic corpus. The second reason is that the AWL and NAWL are classics and landmarks as academic lists in vocabulary and corpus-based lexical studies. Our aim is not to replace the AWL and NAWL, but create a large corpus and scientific dictionary representing research papers from various subject fields without any limitation in disciplines for our goal of discovering and quantifying the meaning of research texts.

The NAWL consists of 963 word families based on a academic corpus of 288 million running words (all words in text without tables' captions, titles and references) \cite{browne2}. The major categories of the corpus where words are extracted are: the Cambridge English Corpus (CEC), oral sources and textbooks. The largest proportion of tokens came from the CEC, about 86\% (over 248 million words). The oral part was taken from the Michigan Corpus of Academic Spoken (MICASE) and the British Academic Spoken English (BASE) corpus. The oral corpora and the corpus of textsbooks are divided into four categories: Arts and Humanities, Life and Medical Sciences, Physical Sciences, and Social Sciences. The NAWL covers 92\% of its corpus when combined with the New General Service List (NGSL) \cite{browne1}. In the list, words are listed by headwords of word families where various inflected forms are contained in. However, we observed that some of headwords in the NAWL indicate the same stemmed word in the LScDC. For instance, two distinct headwords `accumulate' and `accumulation' in the NAWL matched with `accumul' in the LScDC.  To avoid the effect of this difference on our analysis, we applied stemming on words of the NAWL. We used average statistics of different forms as the final statistics for stemmed word. After stemming, the number unique words (lemmas) decreased to 895 in the NAWL, we take these words into account in the comparison study.
 
The NAWL hovewer does not contain much specialised technical terms such as chemical terms and species in biology. To illustrate this, let us explain more details. The Coxhead’s Academic Word List (AWL) created in 2000 -- inspried to create NAWL --, includes 570 word families (from a corpus of 3.5 million words) \cite{coxhead}. According to Coxhead, academic words are supportive of academic text but not central to the topics of the text. The list of AWL contains words that account for approximately 10\% of the total words in the collection of academic texts. The AWL and GSL (General Service List) covers approximately 86\% of total words in academic corpus. Coverage refers to the number of words (tokens, i.e. all forms of words) in texts which are covered by the list of words. By updating this list with an expended and carefully selected corpus of 288 million words (corpus where NAWL is designed), the coverage was improved to 92\% of new corpus when combined with NGSL, with approximately 5\% improvement \cite{browne2,browne1}. Not all words extracted from the corpus is included in the list. In the creation of list, a measure considering the distribution of words over disciplines (Dispersion) was taken into account as well as the frequency (Standard Frequency Index). In LScDC, we include all of words appearing in not less than 10 texts in the corpus -- distributed over 252 subject categories --, we did not apply any procedure to exclude any other words. This explains the difference between the number of lemmas in two lists -- with 895 lemmas of NAWL and 104,223 lemmas in the LScDC. 

The NAWL and the LScDC were actually developed from different corpora and the number of words are quite different, but overall the LScDC contains the much lemmas of NAWL (except only 4 words). In this stage, we did not include the New General Service List (NGSL) as our aim is to evaluate only academic words. In comparison, it must be stressed that there is 103,328 more lemmas in LScDC than the NAWL. Adding the NGSL could result in an increase in the number of the same words in the LScDC and NAWL plus NGSL. 

In the comparison study, we initially investigate dictionaries to see the coverage of NAWL words by LScDC. We will see that there are 891 words that occur in both the LScDC and the NAWL, which means that the overlap between the LScDC and the NAWL is 99.6\%. Four words appearing only in the NAWL are: ``ex", ``pi", ``pardon" and ``applaus". This seems to be the result of differences in types and processing of texts in corpora. It is worth to note that corpus of NAWL includes full texts from academic domain, while the LSC includes abstracts of academic texts. This, for instance, may be the reason why ``pi" does not appear in LSC as it is commonly used by the symbol $\pi$ (pi) in math world and not many articles include formulas in abstract. The other two words ``pardon" and ``applaus" are contained in LScD with low frequencies (5 and 9 respectively); therefore, they are removed in the step of LScDC creation. However, these two words have low rank in the NAWL as well: rank 924 and 956 in the list). Finally, the word ``ex" does not occur in the LScDC due to pre-processing steps applied in the creation of the dictionary. We united some prefixes including ``ex" with the following words (e.g. ex-president is converted to expresident). 

We also present different approaches for comparison to understand what fragment of LScDC contains the NAWL. This is performed by repeatedly searching NAWL words in various subsets of LScDC. Our second focus in dictionary comparison will be the comparison of ranks of words in two dictionaries. In this study, only common words (891 words) are taken into account. Several different methods to compare ranks are considered such as direct comparison of ranks, pairwise comparison of partitions in dictionaries (lists are divided into sub-lists and overlapping words are count in each sub-list), comparison of the top $n$ words, and the comparison of the bottom $n$ words. We will also test similarity of ranks by statistical tests. It is expected to observe that words in two lists are not distributed in the same way as the statistics to order lists are not calculated in the same way. The NAWL considers the dispersion of words over categories, while we simple take the number of documents containing words in the LScDC. All approaches and  results are presented in the section of comparison in detail.   

In this study, we also consider the reproducibility of dictionaries from the LSC and list of texts from other sources to be used by researchers in many other text mining applications. For this purpose, we made R codes for producing the LScD and the LScDC, and instructions for usage of the code available in \cite{code}.

\subsection{The Structure of This Paper \nopunct}  \hspace*{\fill} \\

The paper is organised as follows. Section \ref{method} contains the principles in corpus and dictionary design as well as the text and word representation approaches. In Section \ref{relwork}, we describe some of widely used and well-known analogue corpora and dictionaries. Section \ref{lsc} sets out all pre-processing steps in creation process and the structure of LSC. Similarly, Section \ref{lscdsec} and Section \ref{lscdcsec} present pre-processing steps to build the LScD and the LScDC respectively, and the organisation of dictionaries. In Section \ref{compnawllscdc}, a study of comparison of the LScDC and the NAWL with several approaches is contained. Finally, Section \ref{conclusion} concludes the paper. 

\newpage

\section{Methodology} \label{method}
\subsection{Fundementals of Corpus Design\nopunct} \label{corpora} \hspace*{\fill} \\ 

In linguistics, a text corpus is defined as a large collection of text and they are used by linguists, lexicographers, experts in NLP (Natural Language Processing) and in many other disciplines in order to generate language databases, study  general linguistic features, do statistical analysis or learn linguistic rules.  

Types of corpora vary depending on how they are sampled and designed for specific research goals.  Texts in a corpus are assembled to ensure maximum representativeness of a particular language or language variety. Representativeness refers a sample that includes the complete range of texts in a target population \cite{biber}. Target population is closely related to the scope of the research and respectively sampling. Any selection of text is described as a sample; however, representativeness for a sample depends on the definition of the population that sample is intended to represent and methods of selection of the sample from that population. To define the population, the most important two considerations are: boundary of the population (what texts are included) and the range of genres (what text categories are included) within the population. For instance, Lancaster-Oslo/Bergen Corpus (LOB) is defined as the collection of British English texts that all are published in 1961 in the \textit{British National Bibliography Cumulated Subject Index 1960-1964} for books and all 1961 publications in \textit{Willing's Press Guide 1961} for periodicals and newspapers; distributed across 15 text categories (such as general fiction, romance-love story etc.) \cite{mcenery,johansson}. The target population for the LOB was written British English texts that all are published in 1961 in United Kingdom (boundary)--distributed across 15 text categories (genres). 

The goal of corpus construction is very important for corpus design as it determines the target population. For instance, if the goal of the research is to investigate learners' English, it is reasonable to collect essay of students learning English. However, one who wants to capture the complete range of varieties of English will attempt to collect contemporary British English written texts from a wide variety of different domains. With a given research purpose, a simple broad distinction on corpus types can be done: general corpus and specialised corpus. The criteria for representativeness for these corpora differ from each other by sampling principles. A general corpus contains a broad range of genres with a balance of texts from a wide variety of the language in different domains, while a specialised corpus contains texts from a particular genre or a specific time. For instance, a corpus can be representative for general English language which is an example for general corpus; fiction books or researches in medicine which are examples for specialised corpus. 

Some other considerations in sampling decision are the kinds of texts, the number of texts and the length of text samples as well as sampling techniques. Sampling techniques rely on random selection. Basically, selection can be done by a simple random sampling or stratified random sampling. In basic random sampling, texts having equal chance to be selected in a population are randomly selected. In stratified random sampling, the whole population is divided into smaller groups (e.g. genres) and then each subset is sampled using random selection techniques (with proportionality to the subgroups)\cite{leech}. In the LOB corpus, for example, the population was first divided into 15 categories; and samples were drawn from each category.

 \subsection{Representation of Texts and Words\nopunct} \label{representation} \hspace*{\fill} \\

In order for an effective text processing to be accomplished, one of the most fundamental tasks is to select the most appropriate text representation technique for a particular application of NLP. The quality of any text mining and NLP techniques is strongly dependent on the text representation. It aims to represent texts to enable them to be used in mathematical computations by the machine. 

In general, the most common text representation model in text mining is the Vector Space Model (VSM) \cite{salton2,salton3}. In this model, each text is represented by a numerical vector where its components are taken from the content of the text. Components of the vector denote the features that characterise the text such as words, phrases, paragraphs or a single character etc (\textit{tokens}). Therefore, each text is represented by a collection of words (or words' combinations) and the corpus can be represented by the union of such collections. The most common and simplest way to transform a text into a vector space is to represent them by words from the vocabulary of the text collection. Each text in the collection is thus a feature vector in the vector space. In that case, the dimensions of the vector space is equal to the vocabulary size, and the order of the words in each text is ignored. 

Having the texts represented by vector of words, list of words can be extracted with various statistics such as weight of words for a given text. This representation of the text by a bunch of words is called Bag of Words (BoW) \cite{provost}. In BoW, different word weighting schemes can be used. One simple count for each feature's value (word weight) might be Boolean model. In Boolean model, 1 indicates that the word appears in the text and 0 indicates the word does not appear in the text. This scheme holds only the information about presence or absence of a given word in texts. As an extension of the Boolean model, \textit{TF} (term frequency) shows how many times a word appears in the text \cite{salton4}. In this scheme, the distribution of the word across the collection is not taken into account. However, some words can be more significance than others in the corpus. In that case, \textit{TF} can be multiplied by \textit{IDF} (inverse document frequency), which is defined as the logarithm of the division of the corpus size by the number of documents containing the word. This scheme is called \textit{TF-IDF} \cite{ramos}. In addition, another scheme is to  count the number of appearance of a word in the entire corpus when the corpus is considered as one large document.

Designing the texts by words can be performed in different ways depending on the query. One may want to represent text by all inflected forms of words or stop words. For instance, in the creation of the list of the most widely used conjunctions in a language, removing stop words leads to unreliable results. Therefore, the objection in text representation should be to turn each text into a set of words that supply the task with necessary inputs.  

\subsection{Building a Dictionary from Text Collection \nopunct} \label{buildingdic} \hspace*{\fill} \\ 

In this study, a dictionary is defined as the set of unique words (\textit{lemmas}) extracted from texts in a corpus. In other words, a dictionary is produced based on corpus data. In corpus linguistics, every dictionary is compiled from a particular corpus and the way to establish of a word list must be defined individually for a given purpose. 

Dictionaries differ from one another by the words selected. Several distinctions can be done based on their scopes and purposes. From the overview, the simplest distinction can be observed between general and specialised dictionaries (also refereed as \textit{technical dictionaries}). In specialised dictionaries, words are extracted from a corpus in a single (or multi) specific field(s) and indicate the concepts of the field(s) while general dictionaries contain a complete range of words. Words in specialised dictionaries are called \textit{terms} or \textit{topic-specific words}. In the contrast to terms, a word that has a little lexical meaning is called \textit{function word} in linguistics. Some examples of function words are prepositions, pronouns, determiners etc. In English semantic, non-function words (content words) are words that indicate the content or the meaning of the texts such as nouns, verbs and adjectives. In addition, Coxhead used the notion \textit{supportive} for academic words in AWL \cite{coxhead}. She stated that academic words in AWL are supportive of the academic texts. As supportive, she meant words which are not central to the topic of the text. One would consider words that do not indicate any terminology or specialised technical terms in the subject field. `establish' and `inherent', for example, are two of supportive words according to Coxhead. These are words which authors from most or all academic disciplines tend to use them; the majority of them are also used in general English. She excluded all terminologies such as marine species in Oceanography and function names in Mathematics (e.g. Gaussian). 

To define words and dictionaries, several other distinctions are applied by lexicographers and experts such as  prescriptive or descriptive, dictionaries by language, dictionaries by size, Language-for-Specific-Purposes dictionary (LSP such as medical dictionaries) etc \cite{bowker}. In this study, rather than consider such distinctions, we focus on building a corpus-based dictionary from scientific abstracts written in English. Such dictionary may be considered as \textit{scientific dictionary} giving the guidance to scientific writers on such matters as up-to-date, topic-specific and supportive words of academic texts.   

In the creation of a scientific dictionary from academic texts, two important criteria are: corpus size and the variety of disciplines where texts are categorised into. The more texts are collected, the more representative set of words for specific fields can be gathered. Similarly, the more disciplines where collected texts belong to, the bigger and more comprehensive dictionary can be created. A large and multidisciplinary dictionary with all supportive and topic-specific words of academic texts can also cover to other corpora and be used for any text analysis tasks on them.  

\section{Related Works} \label{relwork}
\subsection{Corpora of English\nopunct} \label{analoguecorpora} \hspace*{\fill} \\ 

There are several freely available corpora for NLP tasks. In this section, we begin by listing some of those well-known corpora developed for English. The earliest corpus in electronic form was developed in 1964 at Brown University, which contains written American English published in 1961 \cite{brown,aston}. \textit{Brown corpus} includes 500 samples of American English text of published works in the United States in 1961. Each text consists of over 2,000 words sampled from 15 text categories, with totally over one million running words. Although today the size of corpus is considered small when comparing recent corpora, it is still widely seen as a landmark publication as a computer readable and general corpus among linguistic researchers. The corpus is similarly designed as LOB which followed the design and sampling practice of the Brown corpus in order to match the Brown corpus for British English \cite{mcenery,johansson}. These two corpus became a model for other national corpora, so-called `Brown Family' \cite{leech2}. In selecting texts for inclusion in the Brown corpus and the LSC, different considerations applied based on the aim of the design of corpora as well as the differences in size of corpora. Brown corpus is sampled from a wide variety of different types of sources such as novels, news, editorials, reviews and many more; while the LSC is sampled from scientific abstracts and proceeding papers.   

\textit{British National Corpus (BNC)} is a monolingual, general corpus of over 4,000 samples of modern spoken and written British English covering English of the later part of 20th century (from 1960 onwards) \cite{rayson,leechgn,bnc}. The latest edition of the BNC is published in 2007. In general, it covers many different styles and varieties of text from various subject fields and genres. The written part of the corpus contains samples from a wide source of text such as: regional and national newspapers, journals, academic books, fiction, letters, school and university essays, other literary text. The spoken component of the corpus is made up of informal conversations recorded by volunteers who were selected from different age, social class and gender, and task-oriented spoken language ranging from formal meetings to radio shows and lectures. The corpus was designed to identify social and generic uses of contemporary British English  with 100 million words \cite{aston}. The major differences between the BNC and the LSC lie in the size of the corpus, in the aim of design (being to capture the full range of varieties of contemporary language use versus to extract scientific ones), in the definition of the populations and in the sampling of corpora in terms of being mixed corpus (spoken and written English) versus written English.  

One other well-known corpus is \textit{Oxford English Corpus (OEC)} which is also used by Oxford lexicographers to construct Oxford English Dictionary (OED), supplied by Oxford University Press \cite{oxford}. The corpus contains of over 10 billion words of 20th and 21st century English from English-speaking countries: the UK, USA, Ireland, Australia, New Zealand, the Caribbean, Canada, India, Singapore and South Africa. It is one of the largest corpus in the world \cite{oxford}. The corpus is mainly drawn from the web with all types of English such as academic journals, literary novels, newspapers, magazines, language of blogs, emails and social media \cite{oxford2}. Another Oxford University Press corpus is \textit{Oxford Corpus of Academic English (OCAE)} contains academic journals and textbooks from four main disciplines: physical sciences, life sciences, social sciences, and humanities with 85 million words included \cite{ocae}.

The \textit{SciCorp} is a corpus of 14 English scientific articles sampled from two disciplines: genetics and computational linguistic, released in 2016 \cite{scicorp}. The corpus includes 61,045 tokens. The population of the corpus being compiled from scientific text is similar to the LSC. However, sampling of SciCorp differs from the LSC as being restricted to two disciplines. Apart from sampling principals and the size of corpus, one other difference of SciCorp from the LSC lies in the type of texts: full-text in SciCorp and abstracts of scientific papers in LSC. 

The \textit{Reuters-21578 corpus} (Reuters-21578 Text Categorization Collection) is a collection of 21,578 news documents used for text categorisation \cite{reuters}. It contains news appeared in 1987 with categories. The main differences between the Reuters corpus and the LSC is genre of texts: Reuters corpus contains texts of news while LSC contains abstracts of scientific publications. The LSC is more than 70 times as large as the Reuters corpus.

The \textit{GENIA corpus} is similar to the LSC in terms of the content of texts, both contain the abstracts of scientific papers \cite{genia}. The GENIA corpus is built by annotating abstracts with keywords ‘(MeSH terms) Human’, ‘Blood Cells’ and ‘Transcription Factors’. 2,000 abstracts are selected for a research objective in Biological and Clinical domains. LSC was created without research area restriction and contains 700 times more abstracts from different research areas.

The \textit{DBpedia abstract corpus} contains 4,415,993 texts of the introductory section of Wikipedia articles, these sections may not necessarily be scientific writing \cite{dbpedia}. As introductory section of Wikipedia articles are not actual abstracts of papers, the average length of documents are different than average length of abstracts: 178 words for the LSC and approximately 524 words for the DBpedia.
\\

\subsection{English Dictionaries\nopunct} \label{analoguedictionary} \hspace*{\fill} \\

One question that can not be easily answered in dictionary design is whether there is an exact count of the number of English words. The major reason for this issue lies in the dynamic nature of languages. It is comonly accepted that languages change rapidly with cultural and technological evalution, and adoption from other languages \cite{steels}. For instance, the \textit{Oxford English Dictionary} has recently added `satoshi' (the smallest unit of a bitcoin), `yeesh' (expressing exasperation, annoyance, disapproval) and `simit' (a type of ring-shaped bread roll originating in Turkey) to its database in 2019 \cite{satoshi,yeesh,simit}. Another consideration on counting the number of words is that what words a dictionary includes. For example, a dictionary would include all technical terms, scientific entries or slang; all of the inflected form of a word (e.g. listen, listening etc.); plurals of words as separate word; or compounds which is made up two words. Therefore, the simple question `what exactly is a word?' turns out to be surprisingly complicated. Some dictionary-makers agree that different versions of words should be counted only once, while some others consider each form as a separate word \cite{crystal}. This means that there may be unlimited number of words in writing and spoken English, which do not appear in any dictionary. 

Although it is not possible to know exact number of words in English, the estimate has been given roughly one million words (ranging from half a million to over two million) -- including names of chemicals and scientific terms-- in vocabulary \cite{webster,harward}. Many of these words are too rarely used, so it is expected that they do not appear in any English dictionary. One of the most well-known and commonly used dictionary \textit{Oxford English Dictionary (OED)} includes over 600,000 words recorded in 20-volumes \cite{oxford3}. The dictionary provides both present days meaning of the words and the history of words from the across of English speaking countries. In addition to the print edition, the dictionary is available online \cite{oxfordonline}. Similar to the OED, the \textit{Webster's Third New International Dictionary} contains over 470,000 entries \cite{webster,webster2}. Another Oxford dictionary so-called \textit{New Oxford Dictionary for Writers and Editors} is built to guide writers, editors, journalists and everyone who works with words \cite{ritter}. It includes 25,000 words and phrases with providing advice on spelling, capitalisation, specialist words and cultural context such as names, mathematical symbols, chemical elements. The \textit{Oxford Learners's Dictionary of Academic English (OLDAE)} is also supplied by Oxford University Press with over 22,000 words based on the OCAE \cite{lae}. The aim of the dictionary is to help students particularly in academic English writing. As an example of specialised dictionary, \textit{Stedman's Medical Dictionary} contains more than 107,000 terms with images (including abbreviations and measurements) in medical references in its 28th edition \cite{stedman,stedman2}. It is designed to provide language of medicine, nursing and health profession to medical students, researchers, physicians and many more medical language users. Finally, we paid attention to the work of AWL and NAWL as they are classics as academic word list \cite{coxhead,browne1}. The AWL includes 570 word families from the collection of written academic texts distributed across the four main disciplines: arts, commerce, law and science. It covers 86\% of the total words in the corpus when combining with GSL. Similarly, NAWL contains 963 word families based on an up-to-date corpus of academic texts. NAWL-NGSL covers 87\% of new corpus. The more detailed explanation is given in the Section \ref{compnawllscdc}.  

Although the estimation of number of words in a language is not a easy task and numbers of words in dictionaries vary differently depending on the content of the dictionary, a corpus-based analysis may give a sight to understand the average number of words for a vocabulary. Let us consider Oxford English Corpus and Oxford English Dictionary with base forms of words (lemmas). It is stated in \cite{archiveoec} that 25\% of all words used in OED is one of lemmas: the, be, to, of, and, a, in, that, have and I. These are the most common 10 lemmas in English. In similar way, the most common 100 and 1,000 lemmas account for  50\% and respectively 75\% of all words used in OEC. To cover 90\% of the corpus, one needs 7,000 lemmas. 95\% of the corpus includes approximately 50,000 lemmas which words in between occur very rarely (e.g. only once every several million words). To cover 99\% of the corpus, we need a vocabulary of over 1 million lemmas. In that case, many words may appear only once or twice in entire corpus (e.g. specialised technical terms), but lemmas will be representative of the whole corpus. To represent notable part of English, 90-95\% of the corpus may be taken as a reasonable number.

\section{Leicester Scientific Corpus (LSC)} \label{lsc}

Leicester Scientific Corpus (LSC) is a collection of abstracts of articles and proceeding papers published in 2014 and indexed by the Web of Science (WoS) database \cite{wos}. Each document contains the text of abstract and the following metadata: title, list of authors, list of categories, list of research areas, and times cited \cite{woscat,wosra,wostc}. The corpus comprises only documents in English. The LSC was collected in July 2018 and contains the number of citations from publication date to July 2018. 

We describe a \textit{document} as the text of abstract with metadata listed above. The total number of documents in LSC is \textbf{1,673,824} \cite{lsc}. All documents in LSC have non-empty abstract, title, categories, research areas and times cited in WoS databases. There are 119 documents with empty authors list, we did not exclude these documents.

\subsection{Corpus Construction \nopunct} \label{corpusprep} \hspace*{\fill} \\

This section describes all steps in order for the LSC to be collected, cleaned and made available to researchers. Data processing consists of four main steps:

\subsubsection{\textbf{Step 1: Collecting the Data}} \label{step1_corpus}

The dataset is downloaded online  by exporting documents as tab-delimited files, so all documents are available online. The data are extracted from Web of Science \cite{wos}. You may not copy or distribute the data in whole or in part without the written consent of Clarivate Analytics\footnote{Use of the LSC is subject to acceptance of request of the link by email. To access the LSC for research purposes, please email to \url{ns433@le.ac.uk} or \url{suzenneslihan@hotmail.com}.}

\subsubsection{\textbf{Step 2: Cleaning the Data from Documents with Empty Abstract or without Category}} \label{step2_corpus}

Not all papers have abstract and categories in the collection. As our research is based on the analysis of abstracts and categories, preliminary detecting and removing inaccurate documents were performed. All documents with empty abstracts and documents without categories are removed. 

\subsubsection{\textbf{Step 3: Identification and Correction of Concatenated Words in Abstracts}} \label{step3_corpus}

Traditionally, abstracts are written in a format of executive summary with one paragraph of continuous writing, which is known as \textit{unstructured abstract}. However, especially medicine-related publications use \textit{structured abstracts}. Such type of abstracts are divided into sections with distinct headings such as introduction, aim, objective, method, result, conclusion etc. 

Used tool for extracting documents leads to concatenated words of section headings with the first word of the section in abstracts. As a result, some of structured abstracts in the LSC require additional process of correction to split such concatenated words. For instance, we observe words such as “ConclusionHigher” and “ConclusionsRT” etc. in the corpus. The detection and identification of concatenated words cannot be totally automated. Human intervention is needed in the identification of possible headings of sections. We note that we only consider concatenated words captured in headings of sections in medicine-related papers as it is not possible to detect all concatenated words without deep knowledge of research areas. Identification of such words is done by sampling of medicine-related publications. The section headings in such abstracts are listed in Table \ref{table:sechead}.

In headings of a section, the words usually start with a capital letter and end with a colon, unless there is typographical error in an electronic material. The words following a heading word (or a colon) also start with a capital letter in structured abstracts. We take these properties into consideration while detecting concatenated words. 

All words including headings in the Table \ref{table:sechead} are detected in the entire corpus, and then words are split into two words. For instance, the word “ConclusionHigher” is split into “Conclusion” and “Higher”. 

\subsubsection{\textbf{Step 4: Extracting (Sub-setting) the Data Based on Lengths of Abstracts}} \label{step4_corpus}

After correction of concatenate words is completed, the lengths of abstracts are calculated. “Length” refers the total number of words in the text, calculated by the same rule as for Microsoft Word “word count”\cite{word}.
An abstract is a short text that is written to capture the interest of a reader of the paper. Thus, abstracts briefly describe and summarise the work and the findings usually in one paragraph of words, but very rarely more than a page. 

According to APA style manual \cite{apa}, an abstract should contain between 150 to 250 words. However, word limits vary from journal to journal. For instance, Journal of Vascular Surgery recommends that “Clinical and basic research studies” must include a structured abstract of 400 words or less” \cite{vas}. 

\begin{figure}[h]
\centering
 \includegraphics[width=1\linewidth]{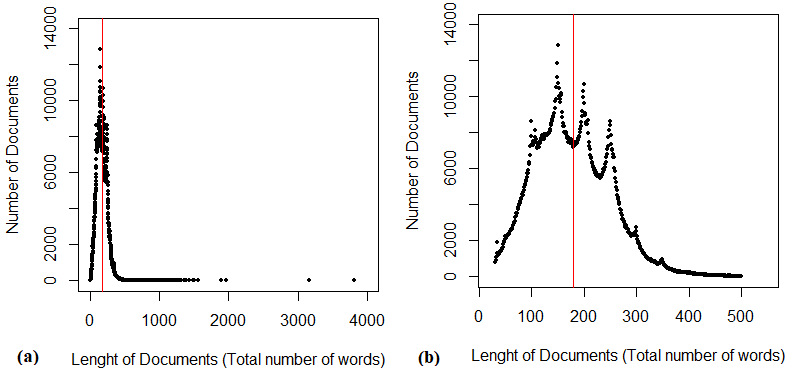}
   \caption{Length distribution of documents (a) before and (b) after removing documents containing less than 30 and more than 500 words (maximum length was 3,805 before removing). The vertical line shows the average length.}
  \label{fig:length}
\end{figure}

In LSC, the length of abstracts varies from 1 to 3,805. We decided to limit length of abstracts from 30 to 500 words in order to study documents with abstracts of typical length ranges and to avoid the effect of the length to the analysis. Documents containing less than 30 and more than 500 words in abstracts are removed. Figure \ref{fig:length} shows the distribution of lengths over documents of LSC before and after removing documents containing less than 30 and more than 500 words.

Four main peaks observed on the length graph: at 100 words, 150 words, 200 words and 250 words. The second peak shows the maximum number of documents, where those observed when the number of words in an abstract is 150. This result is expected as typically word limits range from 150 to 250 words for an abstract. 

After the process of correction and cleaning, the database contains of raw texts of abstracts with title, list of authors, list of categories, list of research areas, and times cited. The total number of documents is 1,673,824 in LSC (see Table \ref{table:correctedtable}).

	\begin{table}[h]
		\centering
		\caption{Number of documents before and after cleaning documents with empty abstracts or without category, and   after removing too short and too long abstracts.}
		\renewcommand\arraystretch{1.4}
  			\begin{tabular}{  l  r  }
	  	 		\hline
	  	 		\rowcolor{pink}
	  	 		\hline
	  	 		\multicolumn{2}{r}{\textbf{\# of Documents}} \\ \hline
	  	Original data	                                                    & 1,727,464	 \\ 
		After cleaning documents with empty abstracts or without category	& 1,681,469  \\ 
		After removing too short and too long abstracts                     & 1,673,824  \\ \hline
			\end{tabular}
	\label{table:correctedtable}
   \end{table}

\subsection{Organisation of the LSC\nopunct} \label{corpusorg} \hspace*{\fill} \\

In LSC, the information is organised with one record on each line and parts of “List of Authors’, “Title”, “Abstract”, “Categories”, “Research Areas”, “Total Times cited”, and “Times cited in Core Collection” is recorded in separated fields \cite{lsc}. Table \ref{docst} demonstrates the structure of a document in LSC. 

The “Categories” field contains the list of the subject categories where the document is assigned to \cite{woscat}. Each document in LSC is assigned to at least 1 and at most 6 categories. There are totally 252 categories in the corpus. The full list of categories are presented in Table \ref{table:categories} and \cite{lsc}.

The “Research Areas” field consists of the list of research areas described as “a subject categorisation scheme” in WoS database \cite{wosra}. Each category is mapped to one research area in the WoS collection. There are totally 151 research areas in the corpus. The full list of research areas is presented in Table \ref{table:researcharea} and \cite{lsc}.

“Total Times Cited” consists the number of times the paper was cited by other items from all databases within Web of Science platform. A paper can appear in multiple databases indexed in Web of Science collection. The citation indexes in WoS are: Web of Science Core Collection, BIOSIS Citation index, Chinese Science Citation Database, Data Citation Index, Russian Science Citation Index and SciELO Citation Index. Duplicate documents across multiple databases is counted only once \cite{wostc}. 

“Times Cited in Core Collection” is the total number of times the paper cited by other papers within the WoS Core Collection. The citation indexes in Core Collection are: Science Citation Index Expanded, Social Sciences Citation Index, Arts and Humanities Citation Index, Conference Proceedings Citation Index–Science, Conference Proceedings Citation Index–Social Sciences and Humanities, Book Citation Index–Science, Book Citation Index–Social Sciences and Humanities, Emerging Sources Citation Index \cite{wostc}.

\section{Leicester Scientific Dictionary (LScD)} \label{lscdsec}

This section presents the pre-processing steps for creating an ordered list of words from the LSC \cite{lsc} and the description of \textit{Leicester Scientific Dictionary (LScD)}. 

LScD is an ordered list of words from texts of abstracts in LSC \cite{lscd}. The dictionary is sorted by the number of documents containing the word in descending order. The dictionary stores 974,238 unique words, where abbreviations of terminologies and words with number are contained in. All words in the dictionary are in stemmed form of words. The LScD contains the following information: unique words in abstracts in the LSC, number of documents containing each word and number of appearance of each word in the entire corpus.

“The number of documents containing a word” is the number of the documents with the corresponding word. A word that appears multiple times in a document is counted once (binary representation for existence). “Number of appearance of a word in the entire corpus” is defined to be the total number of occurrences of a word in the LSC when the corpus is considered as one large document.

All words obtained after pre-processing steps are included in the LScD. The most frequent 20 words (frequency is calculated by the number of documents containing a word) are presented in Table \ref{table:mostfreq} . 

	\begin{table}[h]
		\centering
		\caption{The most frequent 20 words in the LScD.}
		\renewcommand\arraystretch{1.3}
  			\begin{tabular}{ |L{1.6cm} |   R{4cm}  | L{1.6cm}  | R{4cm} | }
	  	 		\hline
	  	 		\rc \textbf{Word} &\rc \textbf{Number of documents containing the word} & \rc \textbf{Word} & \rc \textbf{Number of documents 
containing the word} \tn \hline

 use 	 &   902,033 	& also 	    &   400,642  \\ \hline
 result  &	 812,154 	& present 	&   389,735  \\ \hline
 studi 	 &   723,827    & increas 	&   383,676  \\ \hline
 show 	 &   498,705 	& two 	    &   375,586  \\ \hline
 method  &   491,586    & model     &	372,911  \\ \hline
 effect  &   476,757 	& signific 	&   370,435  \\ \hline
 base 	 &   446,436 	& compar 	&   355,381  \\ \hline
 differ  &   445,739 	& paper 	&   346,514  \\ \hline
 can 	 &   441,512 	& time 	    &   344,817  \\ \hline
 high 	 &   402,737 	& perform 	&   341,547  \\ \hline
			
			\end{tabular}
	\label{table:mostfreq}
	\end{table}

\subsection{Processing the LSC and Building the LScD\nopunct}\label{lscd} \hspace*{\fill} \\

The main challenge of using text data is that it is mess and not concretely structured. This means that a number of steps is needed to be taken to form the LScD. The initial step of building the dictionary is to convert unstructured text (raw corpus) into structured data. Structured data means highly organised and formatted in a way so the information contained can be easily used by data mining algorithms, mostly numerical data in relational databases \cite{blumberg}. There are different ways to pre-process text data and pre-processing steps should be described for each corpus individually. Decision taken and steps of processing for creation of LScD are described below. All steps can be applied for arbitrary list of texts from any source with changes of parameters and also to LSC to reproduce the dictionary.  
\subsubsection{\textbf{Step 1: Text Pre-processing Steps on the Collection of Abstracts}} \label{step1_lscd}
Text pre-processing means to bring the text into a form of analysable for the task. This step is highly important for transferring text from human language to machine analysable format by data mining algorithms. As each task requires different procedures to process the text based on aim of the study, ideal pre-processing procedure of each task should be developed individually. We used    standard pre-processing methods in text processing studies such as tokenization, stop word removal, removal of punctuations and special characters, lowercasing, removal of numbers and stemming as well as two non-standard pre-processing steps: uniting prefixes of words and substitution of words. In this section, we present our approaches to pre-process abstracts of the LSC. 

 \begin{enumerate}
	 \item \textbf{Removing punctuations and special characters:} This is the process of substitution of all non-alphanumeric characters by space. We did not substitute the character “-” in this step, because we need to keep words like “z-score”, “non-payment” and “pre-processing” in order not to lose the actual meaning of such words. A processing of uniting prefixes with words are performed later. 
 	\item \textbf{Lowercasing the text data:} Lowercasing is one of the most effective pre-processing step in text mining problems to avoid considering the same words like “Corpus”, “corpus” and “CORPUS” differently. Entire collection of texts are converted to lowercase. 
	\item \textbf{Uniting prefixes of words:} Prefixes are letters placed before a word to create a new word with different meaning. Words containing prefixes joined with character “-” are united as a word. The list of prefixes united for this research are listed in Table \ref{table:prefix}. The most of prefixes are extracted from \cite{prefx}. We also added commonly used prefixes: “e”, “extra”, “per”, “self” and “ultra”.
	\item \textbf{Substitution of words:} Some of words joined with “-” in the abstracts of the LSC require an additional process of substitution to avoid losing the meaning of the word before removing the character “-”.  Some examples of such words are “z-test”, “well-known” and “chi-square”. These words have been substituted to “ztest”, “wellknown” and “chisquare”. Identification of such words is done by sampling of abstracts from LSC. The full list of such words and decision taken for substitution are presented in Table \ref{table:subs}. 
	\item \textbf{Removing the character “-”:} All remaining character “-” are replaced by space. 
	\item \textbf{Removing numbers:} All digits which are not included in a word are replaced by space. All words that contain digits and letters are kept for this study because alphanumeric characters such as chemical formula might be important for our analysis. Some examples of words with digits are “co2”, “h2o”, “1990s”, “zn2” and “21st”. 
	\item \textbf{Stemming:} Stemming is the process of converting inflected words into their word stem. In this process, multiple forms of a specific word are eliminated and words that have the same base in different grammatical forms are mapped to the same stem. As stemming removes suffixes and reduces the number of words in corpus, this step results in uniting several forms of words with similar meaning into one form and also saving memory space and time \cite{ramya}. For instance, the word “listen” is the word stem for “listens”, “listened”, and “listening”. All words in the LScD are stemmed to their word stem by R package \cite{bouchet}.
	\item \textbf{Stop words removal:} In natural language processing, stop words (including function words) are defined as words that are extreme common but provide little value in a language. Some common stop words in English are “I”, “the”, “a” etc. Such words appear to be of little informative in documents matching as all documents are likely to include them. There is no universal list of stop words. Stop words must be chosen for a given purpose.  In our research, we used “tm” package in R to remove stop words \cite{feinerer}. There are 174 English stop words listed in the package. Full list of stop words in tm package can be found in Table \ref{table:stopwords}. 
 \end{enumerate}

\subsubsection{\textbf{Step 2: Extracting Words from Abstracts}}\label{step2_lscd}

After pre-processing the abstracts of LSC, there are 1,673,824 processed plain texts for further analysis. All unique words in the processed texts are extracted and listed in the LScD.

\newpage
\subsection{Organisation of the LScD\nopunct}\label{lscdorg} \hspace*{\fill} \\

The total number of words in LScD is 974,238. Unique words, the number of documents containing the word and the number of appearance of the word in the entire corpus are recorded on each line in separated fields.

The “Word” field contains unique words from the corpus. All words are in lowercase and their stem form. The list of words is sorted by the number of documents that contain words in descending order. 

“Number of documents containing the word” is the number of documents containing the corresponding word in “Word” field. In this content, binary calculation is used: if a word exists in an abstract then there is a count of 1. If the word appears more than once in a document, the count is still 1. Total number of document containing the word is counted as the sum of 1s in the entire corpus. 

A word can appear many times in the same document. “Number of appearance of a word in the entire corpus”  is computed as the sum of appearance of the word in each document. The field contains how many times a word occurs in the corpus when the corpus is considered as one large document. 

\subsection{Basic Statistics in the LScD\nopunct}\label{lscdstat} \hspace*{\fill} \\

\begin{figure}[hbt]
\centering
 \includegraphics[width=0.65\linewidth]{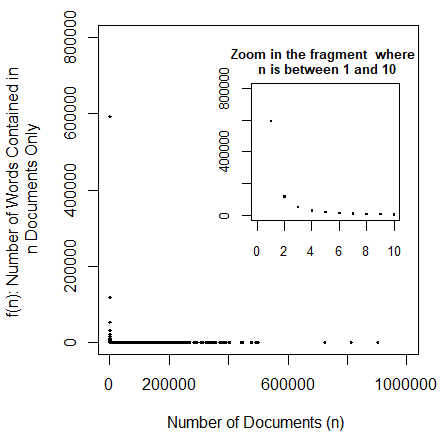}
   \caption{The number of documents (n) versus the number of words contained in n documents only.}
  \label{fig:docswords}
\end{figure}

Before moving on creation of a core dictionary LScDC from LScD, we investigated basic statistics of LScD. The Table \ref{table:freqwords} shows the number of the rarest words over documents, where words appear in at most 20 documents. For instance, there are 592,161 words contained in only 1 document in the corpus. This distribution is also presented for all words in the Figure \ref{fig:docswords}. As expected, very few words occur very often, there is a larger number of mid-frequency words and very many words occur very rare in the collection. This is a typical property of text data and the distribution of words in texts \cite{zipf}. 

\begin{table}[h]
		\centering
		\caption{The number of documents and the number of words contained in the corresponding  number of documents only for those words appearing in at most 20 documents.}
		\renewcommand\arraystretch{1.1}
  			\begin{tabular}{ | m{2cm} |  R{3.2cm}  | L{2cm} | R{3.2cm} |  }
	  	 		\hline
     		\rc \textbf{Number of Documents}	& \rc  \textbf{Number of Words  Contained in the Corresponding Number of Documents Only}   &\rc  \textbf{Number of Documents}	& \rc \textbf{Number of Words  Contained in the Corresponding Number of Documents Only}  \tn \hline
			1  	&  592,161    &  11      &  5,605      \\ \hline
		    2 	&  118,989    &  12	     &  4,912      \\ \hline
		    3  	&  54,193     &  13      &  4,268      \\ \hline
			4	&  32,032     &  14      &  3,689	   \\ \hline
			5	&  21,624     &  15      &  3,385      \\ \hline
			6	&  15,554     &  16      &  2,971      \\ \hline
			7	&  11,877     &  17      &  2,752      \\ \hline
			8	&  9,384      &  18      &  2,522      \\ \hline
			9	&  7,709      &  19      &  2,253      \\ \hline
			10  &  6,492      &  20      &  2,161      \\ \hline
			\end{tabular}
	\label{table:freqwords}
	\end{table}

\subsection{Decision Taken for Rare and Frequent Words\nopunct}\label{lscdrare} \hspace*{\fill} \\

\subsubsection{\textbf{Traditional Approaches to Rare and Frequent Words} \nopunct} \hspace*{\fill} \\\\
In most studies on text classification and information extraction, it is common to discard rare words in order to improve the performance of methods. The idea of the usability of rare and frequent words for discriminating texts dates back to Luhn’s idea \cite{luhn}. He proposed a model to automatically generate the abstract by extracting the most representative sentences among all the sentences in an article. To select those sentences, a measure of information based on an analysis of words in sentences is used. It is assumed that the word occurrence in an article can be used to compile a set of significant word, and the frequency of such significant words within sentence reflects the “significance of sentence” in the text.  According to Luhn, rare and frequent words in a text do not contribute much to the content of the text. Luhn stated that only words between two cut-offs, middle frequency words, can be determined as significant words for the text.   

Besides extraction of significant words in an article, such an analysis can be also applied to the collection of documents to extract the most significant words to discriminate articles across the collection. In other words, significant words can be extracted on a corpus basis rather than a per-document basis. Luhn’s original idea of counting frequencies can be used to provide weighting to words in order to discriminate documents in a collection. Following to this, it is showed that words appearing in low number of documents and words appearing in high number of documents are not good discriminators across the collection in \cite{salton}. They verified Luhn’s conclusion that words with middle frequency are the best discriminator, and words occurring in between 1\% and 90\% of texts have the highest discriminatory power across the collection. Pruning rare and frequent words is followed by many researchers in text categorisation tasks as it is a common belief that they are not good discriminators of classes \cite{torrisi}. 

However, as noted in \cite{yangpedersen} common words (or frequent words) contribute to the text categorisation contrary to a common belief that the removal of frequent  words  improves the performance of information retrieval methods. In their study, stop words are discarded before evaluation of the performances. They also examined another common assumption that rare words are informative and should not be removed; and concluded that words appearing in less than some pre-determined number of documents (up to 90\% or more unique words) can be removed with either an improvement or no loss in the accuracy of text categorisation models. Similar conclusion is obtained for clustering in \cite{rigouste}. They investigate the side effect of vocabulary size to clustering algorithms. The results show that keeping frequent words leads to an improvement on the performance of models in general, while rare words can be removed without loss on the performance. It is also worth to mention that the score that is used to evaluate mutual information is almost 0 when using only words with 1 occurrences in the corpus. Similar results are observed in \cite{wayne} for their scoring measure where rare words skew the distribution of score defined. A minimum threshold of 10 is used in our study to mitigate this issue.

In information retrieval systems, a common belief among researchers is that both frequent and rare words can be important for a specific field. However, the extraction of words from these two classes should be done by using two criteria not one as their distributions are different in specific areas \cite{nugumanova}. Rare words can be topic-specific words and extracted by analysing their co-occurrence in the academic domain. It is stated that a rare word will be a topic-specific word if it is related to a huge number of other possible topic-specific words or words that are considered as informative with a large weight defined \cite{hasan}. In \cite{weeber}, it is reported that most of rare words that are generally discarded in standard information extraction tasks can be topic-specific words in medical abstracts. They stated that even the frequency of 5 is too high for extraction of informative words in medical abstracts but words appearing only once is needed to be removed in information based statistical models \cite{nugumanova,hasan,weeber}.

\subsubsection{\textbf{Characteristics of Words in the LScD and the Decision Taken} \nopunct}\hspace*{\fill} \\\\
In practice, word selection strategy is fundamentally important for different text processing tasks since it determines the space of words that can be obtained from the texts and be effectively used for a specific task. Differences in types of corpora must also be considered as a complementary effect in the selection of words. In order for the differences between rare/frequent words’ importance in two corpora to be explainable, corpora should be comparable by sampling in the same way. For example, it is natural to expect that frequent words of a topic-specialised corpus are different from frequent words appearing in a general corpus where its texts are from a wide variety of different domains. For information extraction problems, frequent words in a topic-specialised corpus are likely to be extracted as content words while such words can be assessed as non-informative in a general corpus. As mentioned before, 5 occurrences of a word in a topic specific corpus (e.g. medical abstracts) may be too high when compared with a general corpus of the same size. However, one can find that words with 5 occurrences are useless for text categorisation tasks due to its score in probabilistic models (e.g. entropy).  

Essentially, rare words fall into two classes: those which are rarely used in the corpus and those which are misspelling. There are several reasons for the first class. It may be because it is used very uniquely like names referring people, places, brands or products. Rare words may also refer infrequent usage or synonyms of words. Similarly, shorten version and abbreviations of words can cause a huge number of rare words. Particularly, those who use corpus containing texts from medical or chemistry domains will tend to see huge number of shorten words, abbreviations and also chemical formulas. The second class of rare words involves words that are misspelled in the writing. Especially, such words is one of the main factor contributing the number of words occurring once in large corpora. As one would expect, a list with all correctly spelled words would not be realistic, especially for a large corpus. In \cite{wilbur}, it is predicted that 38\% of 42,340 words, from a collection of life science abstracts, are misspellings. For both classes of rare words, one needs to be careful about removing them. The decision of cut-off for rare words should be determined individually for each corpus depending on the characteristic of the corpus such as type and size. 

Therefore, a natural question arises: what is the optimal cut for rare words in LSC? A simple initial characterisation is taken into account. As mentioned before, LSC is a collection which texts are from 252 different categories. Two expected consequences of this fact are: the identification of informative rare words for text categorisation by using their co-occurrences with other words of the corpus is not reasonable for our case; and it is very likely to observe words occurring only once in the corpus. The first consequence is caused by the fact that two rare words that used in texts from two well-separated categories will tend to be associated with each other due to co-occurrence of these words with the same subset of other words. In the case that the subset of related words has a large weight in terms of containing informative words but one of rare words is actually not informative, the selection of this rare word will be biased on the other one. When considering the large size of LSC, having a large number of categories has also a side effect: many misspellings and unique names. In fact, approximately 60\% of words appear in only 1 document in LSC (Table \ref{table:freqwords}). Casual observation of words showed that many of them are non-words or not in an appropriate format to use (e.g. misspellings); therefore, they are likely to be non-informative signals (or noise) for algorithms. Some examples of such cases in LSC for randomly selected rare words are presented in Table \ref{table:raresample}. Our basic assumption is that too rare words are not informative for text categorisation, or not effective in the performance of methods.

\begin{table}[h]
		\centering
		\caption{Some of rare words in LScD with the number of documents containing them. The last column shows the description of the word provided by checking the papers containing the word in WoS database, and possible reason why it is rare.}
		\renewcommand\arraystretch{1.3}
  			\begin{tabular}{ | m{2cm} |  R{2cm}  | m{7.5cm}|  }
	  	 		\hline
     		\rc  \textbf{Word} 	&  \rc \textbf{Number of documents containing the word}    &  \rc\textbf{Description of the word and possible reason why it is rare}	\tn  \hline
     		
luhman	   		& 4	 & An author name 																								 \\ \hline
lazerian		& 5	 & An author name																								 \\ \hline
goodluck		& 2	 & A name (President Goodluck Jonathan)																		     \\ \hline
hansel	    	& 5	 & A name (a name in fairy tale Hansel and Gretel and an author’s name)										     \\ \hline
masculina		& 8	 & A marine specie: Appendix Masculina (Latin name)															     \\ \hline
heterocop		& 1	 & A freshwater specie: Heterocope Borealis (Latin name)														 \\ \hline
lunac18(2)		& 1	 & A term in Chemistry                                															 \\ \hline
wr3	        	& 3	 & A term in Agriculture (a water regime)																		 \\ \hline
gausian	    	& 3	 & Misspelling- Gaussian																						 \\ \hline
antilmog		& 1	 & Misspelling in the database: AntiLMOG (correct writing in the paper is anti-MOG. 							 \\ \hline
acetosa	   		& 10 & A plant specie Rumex Acetosa (another usage is ‘sorreal’ appearing in 13 documents)										 	 \\ \hline
ansdic	    	& 1	 & An abbreviation for Ammonium Nitrate and Sodium Salt of Dichloroisocyanuric									 \\ \hline
18cm	    	& 10 & Non-word																									  	 \\ \hline
000009sl		& 1	 & \rr Non-word (from the expression ‘DW=0.000009SL(3.047)’) \tn													\hline
limite	    	& 8	 & French word																									 \\ \hline
resultan	    & 1	 & French word																									 \\ \hline
resultadoscon	& 1	 & Spanish word with error: ResultadosCon (‘resultado’ means result in English and appears 90 times in the LSC)  \\ \hline
          
			\end{tabular}
	\label{table:raresample}
	\end{table}
	
In order to mitigate this issue, we set a minimum cut (10) so that words appearing in less than the cut-off will not be included in further analysis. There is no trivial way to decide the optimal cut. We took decision that the threshold which is not too low or high to be able to keep a reasonable number of words for analysis under the assumption that rare words can be relatively informative and they should not be removed aggressively \cite{yangpedersen}. The criteria, removing rare words to improve the performance of information-based text categorisation methods, is taken into account with an attention to have a noticeable impact on size of dictionary and results. 

\begin{figure}[bt]
\centering
 \includegraphics[width=0.65\linewidth]{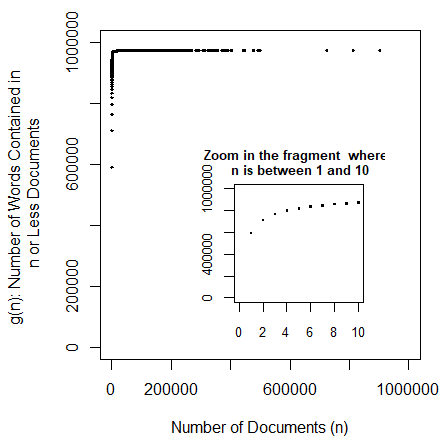}
   \caption{The number of documents (n) versus the number of LScD words contained in n or less documents in the LSC.}
  \label{fig:cumulative}
\end{figure}

\begin{figure}[bt]
\centering
 \includegraphics[width=0.65\linewidth]{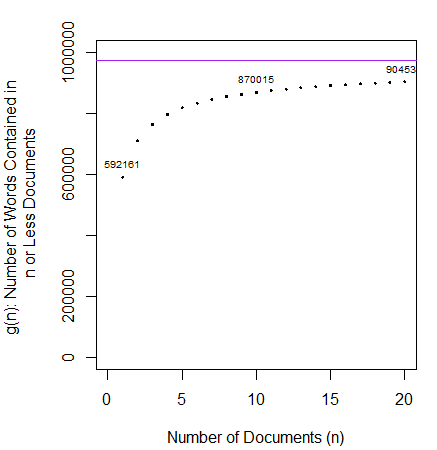}
   \caption{The number of documents (n) versus the number of LScD words contained in n or less documents in the LSC for those words appearing in at most 20 documents. The horizontal line indicates the number of words in the dictionary (974,238).}
  \label{fig:cumulative20}
\end{figure}

The Figure \ref{fig:cumulative} shows the number of words contained in the corresponding or less number of documents. To explore the fragment where words are very rare, we generate an enlarged view on a fragment in the Figure \ref{fig:cumulative20}. For instance, there are 592,161 words containing in only 1 document and 711,150 words containing in 2 or 1 documents.  We can conclude from the figures and Table \ref{table:freqwords} that 870,015 words out of 974,238 words in LScD are contained in 10 or less than 10 documents, thus, it is reasonable to consider such words as non-informative signals in the corpus of 1,673,824 documents and can be removed from the dictionary for further analysis. If such words are removed from the dictionary, the number of words becomes significantly reduced from 974,238 to 104,223. Note that we did not exclude any frequent words in this stage as stop words are already removed in pre-processing steps. 

\begin{figure}[tb]
\centering
\includegraphics[width=1\linewidth]{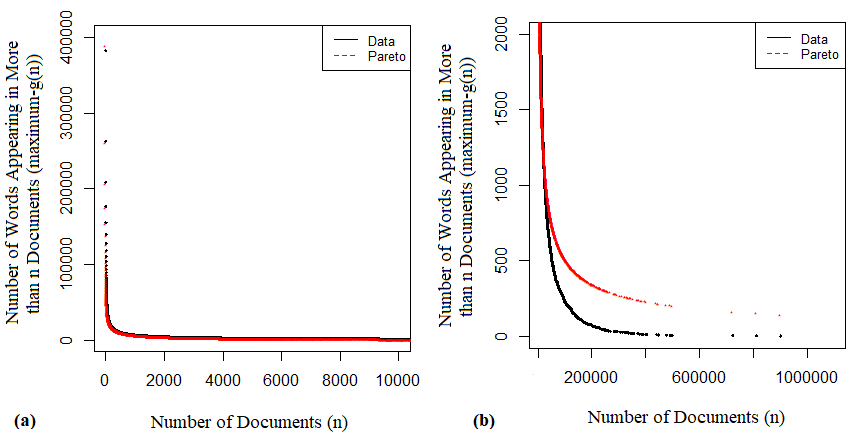}
   \caption{The number of documents ($n$) versus the number of LScD words appearing in more than $n$ documents in the LSC for (a) $n <10,000$ and (b) $n \geq 10,000$. The y-axis is calculated by normalising $g(n)$ to the maximum (maximum- $g(n)$), where $g(n)$ is the number of words contained in n or less documents. The black points are the data; the red points are the fitted Pareto distribution.}
  \label{fig:normalisedfrag}
\end{figure}

\begin{figure}[h]
\centering
\includegraphics[width=1\linewidth]{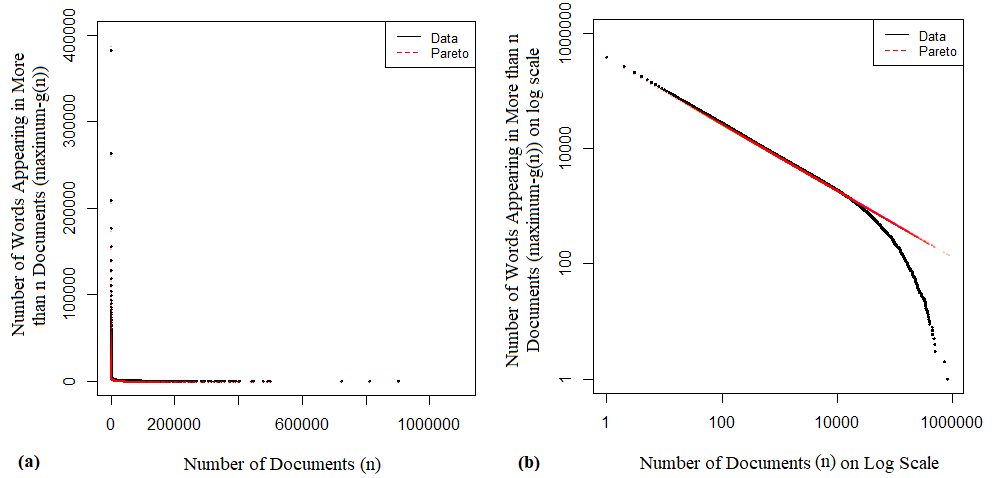}
   \caption{(a) The number of documents ($n$) versus the number of LScD words appearing in more than $n$ documents in the LSC for whole data  (b) The same plot on logarithmic scales. The y-axis is calculated by normalising $g(n)$ to the maximum (maximum- $g(n)$), where $g(n)$ is the number of words contained in n or less documents. The black points are the data; the red points are the fitted Pareto distribution. In (b), the slope of the line is -0.5752.}
  \label{fig:normalised}
\end{figure}

Figure \ref{fig:normalisedfrag} and \ref{fig:normalised} present the normalised number of words contained in the corresponding ($n$) or less number of documents. The data are normalised using (maximum-number of words) on y-axis. This means that the plot shows the number of words appearing in more than $n$ documents against these numbers of documents. We observe more or less a negative relationship on a logarithmic scale. However, a remarkable pattern emerges: the plot does not follow a single straight line which fits the data. Instead, the plot starts with a linear behaviour, and a curve is observable in the right tail of the distribution, which follows a Pareto distribution behaviour. Pareto orginally purposed that the number of people with incomes higher than a certain limit follows a power law \cite{pareto1,pareto2,pareto3,geerolf,newman}. The Pareto principle is also known as 80-20 rule. The rule states that 80\% of people's income is held by the top 20\% of income recipients in the society. Such characteristic of the distribution is also very typical property for the distribution of words over documents in text data. Under Pareto principle, the number of words appearing in more than $n$ documents can be modelled as a power law:

\begin{equation} \label{eq:1}
 N_x=\dfrac{\beta}{{x}^{\alpha}}
\end{equation}
where $ N_x $ is the number of words, $ x $ is a certain documents limit and $\alpha$ and $\beta$ are constants. 

A more general description of the Pareto principle is stated by Pareto distribution. Pareto distribution is a two parameter distribution to fit the trend that a large portion of data is held by a small fraction in the tails of distribution (heavy-tailed distribution) \cite{pickands}. The distribution is characterised by a shape parameter $\alpha$ and a location (scale) parameter $x_m$. The \textit{tail function} and the cumulative distribution function of a Pareto random variable $X$ are given by \cite{hardy,kleiber}:   

\[ P(X>x)= \begin{cases} 
     (\dfrac{x_m}{x})^{\alpha} & x\geq x_m \\
      1 & x< x_m \\
   \end{cases}
\]
and 
\[ F(X)= \begin{cases} 
     1-(\dfrac{x_m}{x})^{\alpha} & x\geq x_m \\
      0 & x< x_m \\
   \end{cases}
\]
where $x_m$ is the (necessarily positive) minimum value of $X$ (the lower bound of the data). The density function is defined as 
\[ f_X(x)= \begin{cases} 
     \dfrac{\alpha x_m^\alpha}{x^{\alpha+1}} & x\geq x_m \\
      0 & x< x_m \\
   \end{cases}
\]

For $ 0< \alpha \leq 1$, the distribution is heavy-tailed and the right tail becomes heavier as $\alpha$ decreases .

In Figure \ref{fig:normalised}, power-law behaviour in the upper tail is well documented. The Pareto distribution (Equation \ref{eq:1}) is fitted to the data and resulting graphs are also shown in Figure \ref{fig:normalisedfrag}. Table \ref{table:parameterspareto} presents the estimated parameters and the mean squared error (MSE). 

\begin{table}[h]
		\centering
		\caption{Estimated parameters of Pareto distribution and the Mean Squared Error (MSE) for LScD.}
		\renewcommand\arraystretch{1.3}
  			\begin{tabular}{ | C{1.5cm} |  C{1.5cm}  | C{1.5cm}|  }
	  	 		\hline
     		\rc  \textbf{$ \alpha $} 	&  \rc  \textbf{$ \beta $}    &  \rc  \textbf{MSE}  \tn \hline
     		     		0.5752 &388,756 &25188 
 \\ \hline
          \end{tabular}
	\label{table:parameterspareto}
	\end{table}

If the logarithm of the number of words appearing in more than a certain number of documents is plotted  against the logarithm of these numbers of documents, a straight line (see Figure \ref{fig:normalised} (b)), where the slope is $\alpha$, is obtained. $\alpha$ is also known as \textit{Pareto index}.

Due to the characteristic of the data, the log-log plot presents a noisy and diffused behaviour in the upper tail.  This is actually because 81 observations fall in the interval 1-100 on y-axis, while there are 5,453 observations lying in the interval 10,000-1,000,000. However, on the logarithmic scale, the size of intervals 1-100 and 10,000-1,000,000 are the same, this leads to a diffusion in the tail. From a heuristic point of view, the plot suggests that there are three subset of words in the collection: too rare words, mid-frequent words and frequent words. A straight down-sloping line covers words the largest part of the list, in which words are not too rare and frequent. It is not actually surprising as words occurring in a few or almost all documents tend to be more evenly diffused across the corpus.

\section{Leicester Scientific Dictionary-Core (LScDC)} \label{lscdcsec}

Leicester Scientific Dictionary-Core (LScDC) is an ordered sub-list from existing LScD \cite{lscdc}. There are 104,223 unique words (lemmas) in the LScDC. To build the LScDC, we decided the following process on LScD: removing words that appear in not greater than 10 documents ($ \leq 10 $). As mentioned before, such words do not contribute much to discrimination of texts as they appear in less than 0.01\% of documents. Ignoring these words has the advantages on the reducing the size of words for applications of text mining algorithms. The core dictionary is also sorted by the number of documents as in LScD.  

Table \ref{table:beforeafter} summarizes the number of words before and after removal. 870,015 words are removed from the LScD, that is, around 89\% of words are removed. After removing such words, we also re-check the number of words in each document to affirm that all abstracts have at least 3 words. We note that in this stage “the number of words in an abstract” does not indicate the length of the abstract but the number of unique content words from the LScDC. After removing 870,015 words from the pre-processed abstracts, all documents have at least 3 unique words. None of documents are removed in this stage. 

\begin{table}[tb]
		\centering
		\caption{Number of words before and after removing words 
appearing in not greater than 10 documents in the LSC.}
		\renewcommand\arraystretch{1.3}
  			\begin{tabular}{ | l |   r | }
	  	 		\hline
	  		 &   \textbf{Number of Words} 	  \\ \hline
LScD  		 &	 974,238 	  \\ \hline
LScDC 	 	 &   104,223      \\ \hline
			
			\end{tabular}
	\label{table:beforeafter}
	\end{table}
	
\subsection{Organisation of the LScDC \nopunct}\label{lscdcorg} \hspace*{\fill} \\
	
In the LScDC, unique stemmed words, the number of documents containing the word and the number of appearance of the word in the entire corpus are recorded on each line in separated fields in the same way as for the LScD \cite{lscd,lscdc}.
	
\subsection{Chracteristics of Words in the LScDC \nopunct}\label{lscdcwords} \hspace*{\fill} \\

\begin{figure}[b]
\centering
 \includegraphics[width=0.6\linewidth]{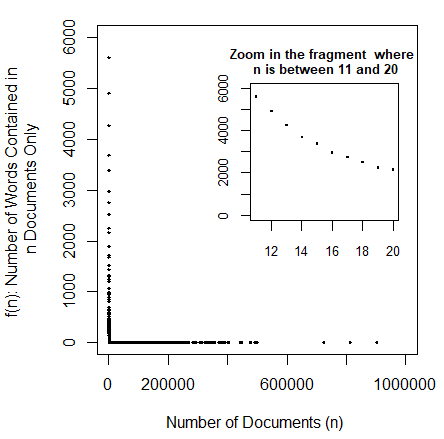}
   \caption{The number of documents (n) versus the number of LScDC words contained in n documents only after cleaning words appearing in not greater than 10 ($ \leq10 $) documents.}
  \label{fig:docswordsreduced}
\end{figure}

After cleaning words appearing in not greater than 10 documents, the distribution of words over documents is presented in Figure \ref{fig:docswordsreduced}. As one can expect, we observe the same behaviour here that very few words occur very often, very many words occur very rare in the collection. 

\begin{figure}[h]
\centering
 \includegraphics[width=0.6\linewidth]{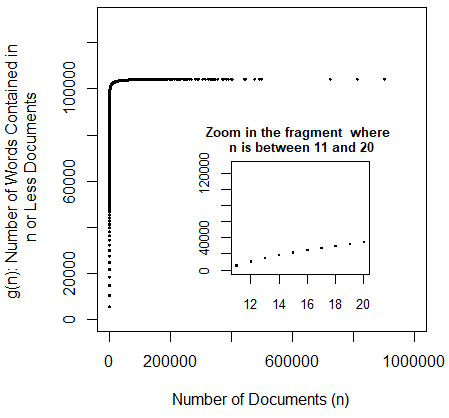}
   \caption{The number of documents (n) versus the number of LScDC words contained in n or less documents in the LSC after cleaning words appearing in not greater than 10 ($ \leq10 $) documents.}
  \label{fig:cumulreduced}
\end{figure}

\begin{figure}[h]
\centering
 \includegraphics[width=0.6\linewidth]{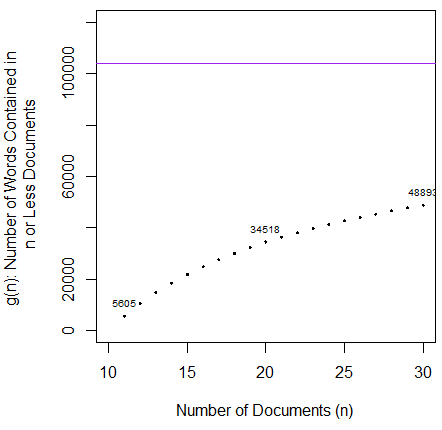}
   \caption{The number of documents (n) versus the number of LScDC words contained in n or less documents in the LSC (for those words appearing in at most 30 documents) after cleaning words appearing in not greater than 10 ($ \leq10 $) documents. The horizontal line indicates the total number of words in the dictionary (104,223).}
  \label{fig:cumul20reduced}
\end{figure}

The Figure \ref{fig:cumulreduced} and Figure \ref{fig:cumul20reduced} show the number of words contained in the corresponding or less number of documents with and without rescaling the x-axis. We can conclude that approximately half of words occur in less than 30 documents.

\begin{figure}[h]
\centering
 \includegraphics[width=1\linewidth]{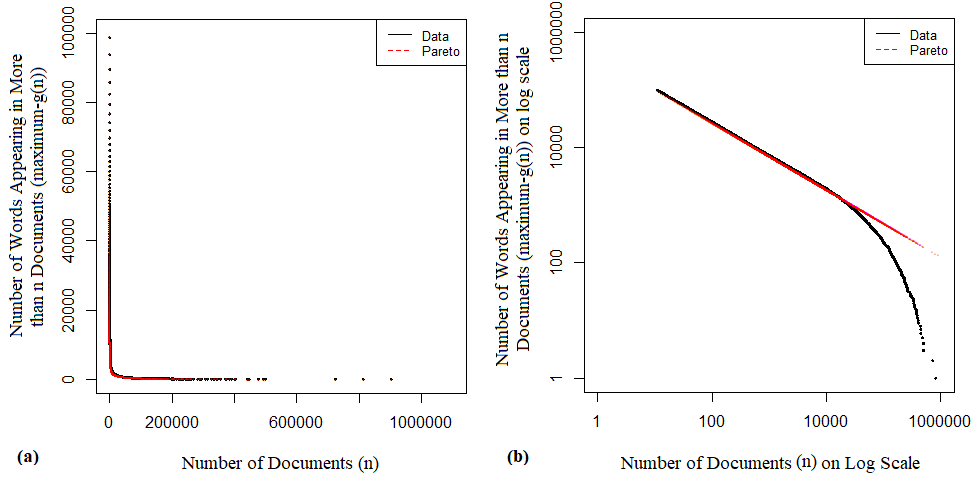}
   \caption{(a) The number of documents ($n$) versus the number of LScDC words appearing in more than $n$ documents in the LSC for whole data  (b) The same plot on logarithmic scales. The y-axis is calculated by normalising $g(n)$ to the maximum (maximum- $g(n)$), where $g(n)$ is the number of words contained in n or less documents. The black points are the data; the red points are the fitted Pareto distribution. In (b), the slope of the line is -0.5797.}
  \label{fig:normreduced}
\end{figure}

Figure \ref{fig:normreduced} demonstrates the normalised number of words contained in the corresponding or less number of documents after removing words appearing in not greater than 10 documents. The data are normalised using (maximum-number of words) on y-axis as in Figure \ref{fig:normalised}. As expected, noisy behaviour in the lower tail is avoided. A downward linear trend is observable at the beginning and a curve is present in the upper tail. From a heuristic point of view, words can be group into two subsets: mid-frequent words and frequent words.

The plots in Figure \ref{fig:normreduced} reveals power-law behaviour (Pareto distribution) in upper tail of documents distribution, but apparently not for the lower tail as expected. The estimated parameters by fitting the power-law (Equation \ref{eq:1}) to the data is presented in Table \ref{table:parameterspareto2}.

\begin{table}[h]
		\centering
		\caption{Estimated parameters of Pareto distribution and the Mean Squared Error (MSE) for LScDC.}
		\renewcommand\arraystretch{1.3}
  			\begin{tabular}{ | C{1.5cm} |  C{1.5cm}  | C{1.5cm}|  }
	  	 		\hline
     		\rc  \textbf{$ \alpha $} 	&  \rc  \textbf{$ \beta $}    &  \rc  \textbf{MSE}  \tn \hline
     		     		0.5796 &397,707 &10737 
 \\ \hline
          \end{tabular}
	\label{table:parameterspareto2}
	\end{table}

\section{A Comparison of LScDC and NAWL} \label{compnawllscdc}

This section provides a comprehensive study of comparison of the NAWL and the LScDC. Several different approaches are taken into account based on direct comparison of words and comparison of ranks of words in two dictionaries.   
\subsection{Academic Word List (AWL) and New Academic Word List (NAWL) \nopunct}\label{nawl} \hspace*{\fill} \\

Academic word list (AWL) is developed from a written academic corpus with 3.5 million running words \cite{coxhead}. The corpus is gathered from four discipline specific sub-corpora: arts, commerce, law and science with a seven sub-disciplines for each (see Table \ref{table:awlstr}). Each sub-corpora has approximately 875,000 running words. The list of words is collected from a total of 414 academic texts in the form of textbooks, articles, book chapters, and laboratory manuals. 

	\begin{table}[h]
		\centering
		\caption{Corpus structure of the AWL.}
		\renewcommand\arraystretch{1.3}
  			\begin{tabular}{ |  l | l | l | l |  }
	  	 	  \hline
	  	 	  \multicolumn{4}{|c|}{\textbf{Discipline}} \\ \hline
 	\textbf{Arts} 	& \textbf{Commerce} 		& 	\textbf{Law} 			& 	\textbf{Science}   \\ \hline
	Education		&	Accounting				&	Constitutional			&	Biology 		   \\
	History			&	Economics				&	Criminal				&	Chemistry 		   \\
	Linguistics		&	Finance					&	Family and medicolegal	&	Computer science   \\
	Philosophy		&	Industrial relations	&	International			&	Geography 		   \\
	Politics		&	Management				&	Pure commercial			&	Geology 		   \\
	Psychology		&	Marketing				&	Quasi-commercial		&	Mathematics        \\
	Sociology		&	Public policy			&	Rights and remedies		&	Physics  		   \\ \hline
	122 texts 		&	107 texts 				& 	72 texts				& 	113 texts          \\ \hline
	883,214 words	& 	879,547 words			& 	874,723 words			& 	875,846 words      \\ \hline
		
			\end{tabular}
	\label{table:awlstr}
	\end{table}

A \textit{word family} is defined as the collection of words that appears in various form of the same word (e.g. indicate and indication are in the same family). To select words, three rules are taken into account:

\begin{itemize}
\item Specialised occurrence: Academic list does not contain the General Service List (GSL) published by West \cite{west}, defined as the first 2,000 frequent words of English.

\item Range: The number of appearance of a family member has to be at least 10 in each of main discipline, and 15 or more in 28 sub-disciplines.

\item Frequency: The number of appearance of a family member has to be at least 100 in the academic corpus. Frequency is the secondary criteria for range.
\end{itemize}

AWL includes 570 word families. It covers 10\% of the total words in academic texts. In addition, words in West’s GSL and words in AWL together (GSL/AWL) cover approximately 86\% of total words in academic corpus. In Coxhead words, “Academic words (e.g. substitute, underline, establish, inherent) are not highly salient in academic texts, as they are supportive of but not central to the topics of the texts in which they occur.” 

The New Academic Word List (NAWL) is then created by \cite{browne1} based on an updated and expanded academic corpus of 288 million words with modified lexemes. The corpus, which NAWL is created from, includes Cambridge English Corpus (CEC), oral academic discourse (Michigan Corpus of Academic Spoken MICASE and British Academic Spoken English BASE), and textbooks (Corpus of 100s top-selling academic textbooks). From CEC, the frequency generated word list is used as one group. This made up the largest proportion of total tokens, about 86\% (over 248 million words). The oral corpora and the corpus of textbooks are divided into four main categories: Arts and Humanities (AH), Life and Medical Sciences (LS), Physical Sciences (PS), and Social Sciences (SS). The number of tokens for each group in the corpus is presented in Table \ref{table:nawlcat}.  The list is developed by the conjunction with New General Service List (NGSL) as in Coxhead’s GSL-AWL. NGSL is also created by \cite{browne1}, which is based on 273 million words from CEC academic. 

\begin{table}[h]
		\centering
		\caption{Corpus Structure of the NAWL.}
			\begin{tabular}{ | L{3cm} | L{4cm} | R{3cm}|  }
	  	 	  \hline
	  	 	  \multicolumn{2}{|C{7cm}|}{\textbf{Source}} 				& \multicolumn{1}{C{3cm}|}{\textbf{\# of tokens}} 			\\ \hline
	  	 	  \multicolumn{2}{|L{8cm}|}{Cambridge English Corpus (CEC)} & \multicolumn{1}{R{3cm}|} {248,666,554}                   \\ \hline
 			  											 				& Arts and Humanities 				& 	803,113 			\\ 
 			   \multicolumn{1}{|L{3cm}|}{Oral Discourse}				& Life and Medical Sciences			&	749,610			 	\\
 			       														& Physical Sciences 				&	 686,926 			\\ 
 			       														& Social Sciences 					&	852,990				\\ \hline
 			       		 
 			  										 					& Arts and Humanities 				& 	6,082,267 			\\ 
 			   \multicolumn{1}{|L{3cm}|}{Textbooks}						& Life and Medical Sciences 		&	16,822,357			\\
 			       														& Physical Sciences 				&	4,467,629 			\\ 
 			       														& Social Sciences 					&	9,044,779			\\ \hline
 			       		 
 			  \end{tabular}
	\label{table:nawlcat}
	\end{table}

NAWL contains of 963 word families. While combined GSL/AWL covers approximately 87\% of the new corpus, the NAWL covers 92\% of the corpus when combined with NGSL. Therefore, NAWL gives an improvement in coverage, with about 5\% more coverage \cite{browne2}. 

In the published list of academic words (NAWL), the authors computed the statistics SFI (Standard Frequency Index), U (Estimated Word Frequency per Million) and D (Dispersion) to describe the number of occurrence of the words and the distribution of words in their corpus. To illustrate the information given in the list, we present Table \ref{table:nawlsfi} that shows 10 words with statistics in the NAWL, ordered by SFI values \cite{carroll1,carroll2, breland}. 

	\begin{table}[h]
		\centering
		\caption{Sample words with highest SFIs from NAWL.}
		\renewcommand\arraystretch{1.3}
  			\begin{tabular}{ |  l | r | r | r |  }
	  	 	  \hline
	\multicolumn{1}{|c|}{\textbf{Word}} & \multicolumn{1}{c|}{\textbf{SFI}} & \multicolumn{1}{c|}{\textbf{U}} &  \multicolumn{1}{c|}{\textbf{D}} \\ \hline
 	repertoire		&	72.452		&	1759	&	0.5923 		   \\
	obtain			&	66.519		&	449		&	0.7531 		   \\
	distribution	&	65.665		&	369		&	0.6863   	   \\
	parameter		&	64.369		&	273		&	0.6943 		   \\
	aspect			&	64.190		&	262		&	0.9385 		   \\
	dynamic			&	63.506		&	224		&	0.8548         \\
	impact			&	63.491		&	223		&	0.9426  	   \\ 
	domain 			&	63.467 		& 	222		& 	0.8276         \\ 
	publish			& 	62.897		& 	195		& 	0.9039         \\ 
	denote			& 	62.571		& 	181		& 	0.7035         \\ \hline	
			\end{tabular}
	\label{table:nawlsfi}
	\end{table}

D shows the uniformity of frequency of the word in subject categories of NAWL in a 0-1 scale: 0 means that the a word (all forms) appears in a single category, 1 means that frequencies are distributed over all categories proportianally to the total number of words (all inflected forms of words) in a category. U is the estimated frequency per million. It is derived form the frequency of the word in the corpus with an adjustment for D. SFI indicates frequency derived from U in a 0-100 scale. Higher scores of SFI show greater frequency \cite{hedgcock}. A word family with SFI=90 occurs once in every 10 tokens (all words with different inflected forms in the corpus); a word with SFI=80 occurs once in every 100 tokens \cite{carroll1,carroll2}. 

\subsection{Difference Between the Principles in Preparetion of the LScDC and the NAWL\nopunct}\label{preperation} \hspace*{\fill} \\

Both the NAWL and the LScDC are actually made up of academic texts distributed over multiple categories for building academic lists of words. In this manner, two lists seem similar. However, more detailed analysis shows that they differ one another in many respects such as types of texts where words are extracted (e.g. full-text or a part of the text), kind of words included, dictionary size and the statistics used to extract words.

Let us begin with corpora where two dictionaries are created. An obvious difference of corpora lies in the types of texts. As types of texts, we meant the NAWL having extracted from full-texts from academic domains and the LScDC having extracted from abstracts of articles. This is actually an important difference as there is side effect of word limit for an abstract such as the frequency of a word and the vocabulary used. In this case, it is likely to observe changes in the statistics calculated for each word and respectively the ranks of words. The change in statistics may lead to select different words as word selection in NAWL is based on frequency and range. One other difference between two corpora is that NAWL contains oral academic discourse as well as written texts while LSC includes only written academic English. This may have influence on the words listed as spoken and written English are often different in terms of vocabulary used.     

It is worth to stress that the calculation of statistics for words in the NAWL and the LScDC are different. In the NAWL, words are selected based on SFI derived from frequency. The dispersion (D) of words over categories is calculated to adjust frequency in SFI calculation. However, in LScDC words are simply sorted by the number of documents containing words. The dispersion of words and SFI are both taken into account to select words in NAWL, not all words appearing in the corpus are included in the NAWL. This difference leads to firstly difference in ranking of common words in both dictionaries, secondly kinds of words and words selected and respectively the size of the dictionaries.

One of the major differences lies in the kind of words. According to the Coxhead, words in AWL are supportive of the academic text but not central to the topics of the text \cite{coxhead}. Words in AWL account for approximately 10\% of the total words in the collection of academic texts. The AWL and GSL (general service list) together cover approximately 86\% of total words in academic corpus. By updating this list with an expended corpus of 288 million words (NAWL), the coverage was improved to 92\% of new corpus when combined with NGSL (New General Service Words), with approximately 5\% improvement. By a casual observation of the NAWL, one can see the same property for words in the NAWL. Words in NAWL are not much specialised technical terms such as names of chemicals or names. In contrary, LScDC contains both supportive and topic-specific words such as mathematical terms, chemical elements, names, biological species and many more. As or aim is to quantifying meaning of research texts, we kept such words in LScDC. 

Such differences in word selection also effected the size of dictionaries. As expected, the LScDC is much more larger than the NAWL, namely 963 word families in the NAWL and 104,223 lemmas in the LScDC.    

\subsection{Comparison of the LScDC and the NAWL  \nopunct}\label{comp} \hspace*{\fill} \\

This section describes a study of comparison of the LScDC \cite{lscdc} and the NAWL. Our primary focus is on obtaining the coverage of NAWL by LScDC, and on analysing how the rank of words in both dictionary are related.

\subsubsection{Coverage of the NAWL by the LScDC \nopunct}\label{coverage} \hspace*{\fill} \\

One feature of NAWL is that words are listed by headwords of word families from combination of their derived forms.  When comparing with LScDC, headwords with different inflected forms indicate the same stemmed word in LScDC (see Table \ref{table:headwords}). In order to examine the agreement between NAWL and LScDC, we processed stemming to headwords in NAWL. This process returns various forms of each headword into a common root as in LScDC. After stemming, words in NAWL are eliminated, with a decrease number from 963 to 895. Note that as SFIs of two headwords, having actually the same root, are different, we used the average of SFIs for unique stemmed words.

	\begin{table}[h]
		\centering
		\caption{Headwords and inflected forms in the NAWL, and stems of the headwords in the LScDC.}
		\renewcommand\arraystretch{1.3}
  			\begin{tabular}{ |  m{3cm} | m{5.5cm} | m{3cm} |   }
	  	 	  \hline
	\rc \textbf{Headword in NAWL} & \rc \textbf{Inflected Forms in NAWL} & 	\rc \textbf{Stemmed Headword in LScDC } \tn \hline
 	accumulate		 		&	accumulates, accumulated, accumulating, accumulatings  &	accumul	 	 \\ \hline	
	accumulation			&	accumulations										   &	accumul		 \\ \hline	
	acid					&	acids												   &	acid		 \\ \hline	
	acidic					&	acidics												   &	acid		 \\ \hline	
			\end{tabular}
	\label{table:headwords}
	\end{table}

For purpose of comparison of dictionaries, stemmed words are used. Table \ref{table:coverage} illustrates the comparison of dictionaries by showing the coverage of the NAWL words by the LScDC. The overlap between the LScDC and the NAWL is 99.6\%, with 891 word occurring in both. This means 4 words occurring only in NAWL: “ex”, “pi”, “pardon” and “applaus”. The lower coverage of the dictionary seems to be the result of differences in types and processing of texts in corpora. The corpus of NAWL includes full texts from academic domain \cite{coxhead}, while LSC is made up abstracts of texts in LSC. 

\begin{table}[h]
		\centering
		\caption{Coverage of the NAWL by the LScDC.}
		\renewcommand\arraystretch{1.3}
  			\begin{tabular}{ |  C{3cm} | C{3cm} | C{2.5cm} | C{2.5cm} |    }
	  	 	  \hline
	\rc \textbf{Number of Words in NAWL} & \rc \textbf{Number of Words in NAWL (after stemming)} & 	\rc \textbf{Coverage of NAWL by LScDC (\#)} & 	\rc \textbf{Coverage of NAWL by LScDC (\%)} \tn \hline
 	 963		 &	 895  &	 891 &   99.6\%	 \\\hline	
		
			\end{tabular}
	\label{table:coverage}
	\end{table}

The reason why “pi” does not occur in LScDC lies in the nature of abstracts and also in the usage of this word in articles. It is commonly used by the symbol $\pi  $ (pi) in the math world, and not many articles include formulas in abstracts.  Uniting prefixes with the following words is the reason that the word “ex” does not occur in LScDC. For instance, words such as ex-president and ex-wife are converted to expresident and exwife in pre-processing step. The other two words “pardon” and “applaus” are not included in LScDC. However, they occurred in LScD before removing words that appear in not greater than 10 documents, with very low occurrences in documents (5 and 9 respectively). Similarly, these two words have low ranks on the NAWL: rank 924 and rank 956 in the list.

\begin{figure}[h]
\centering
 \includegraphics[width=0.6\linewidth]{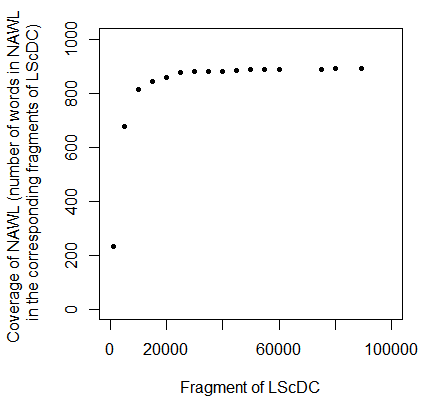}
   \caption{Coverage of the NAWL by the fragments of LScDC.}
  \label{fig:coverage1}
\end{figure}

We also evaluated different comparison scheme that is focused on a subtly different goal: to give an understanding about what fragment of LScDC contains the NAWL. This analysis performs a search of NAWL words over a specific subset of our rank ordered dictionary, repeatedly searching NAWL words in various subsets of the dictionary. Table \ref{table:coverage2} and Figure \ref{fig:coverage1} show the coverage of NAWL in particular fragments of LScDC. From this perspective, we see that NAWL is covered in the first 89,351 (85.7\% of all words) words of LScDC, where the frequency of 89,351th word is 14. Observe that when doubling the number of words from 40,000 to 80,000 there are only 8 more words found in LScDC. This means the majority of NAWL is contained in the first 38.4\% of LScDC. The number of documents containing 40,000th and 80,000th words are 16 and 53 in the LSC. It is remarkable that in 10,000 words, the coverage of the NAWL is 90.9\%, with a frequency of 572 in LSC. This may be considered that the NAWL is representative of our 10,000 words (9.6\% of LScDC). This partly supports that wide range of LScDC is constructed by more specific terminologies of academic disciplines. This is explainable given the variety of texts’ categories in corpus, differences in selection methods of words and the fact that abstracts have slightly different writing structure and words. 

	\begin{table}[h]
		\centering
		\caption{Coverage of the NAWL by the fragments of LScDC. The last column presents 
words of NAWL which are found between two fragments in LScDC.}
		\renewcommand\arraystretch{1.3}
  			\begin{tabular}{ |  R{1.5cm} | R{1.5cm} | R{1.5cm} | R{1.5cm} | L{5cm} |   }
	  	 	\hline
	 \multicolumn{2}{|C{3cm}|}{\textbf{Fragment of LScDC}} & \multicolumn{2}{C{3cm}|}{\textbf{Coverage of NAWL by LScDC}} & \multicolumn{1}{C{5cm}|}{\textbf{Words added between two fragment}} \\ \hline
 	\rc \textbf{\#} & \rc	\textbf{\%}	& \rc  \textbf{\#}  & \rc \textbf{\%}	& \rc  \textbf{Words}   \tn \hline	
		
		1,000 	&	1.0\%	&	231	&	25.8\%	& 											\\\hline
		5,000 	&	4.8\%	&	678	&	75.8\%	&											\\\hline
		10,000	&	9.6\%	&	814	&	90.9\%	&											\\\hline
		15,000	&	14.4\%	&	845	&	94.4\%	&										    \\\hline
		20,000	&	19.2\%	&	860	&	96.1\%	&											\\\hline
		25,000	&	24.0\%	&	877	&	98.0\%	&											\\\hline
		30,000	&	28.8\%	&	879	&	98.2\%	& bizarr, terribl							\\\hline
		35,000	&	33.6\%	&	882	&	98.5\%	& comma,sneez,jazz							\\\hline
		40,000	&	38.4\%	&	882	&	98.5\%	&											\\\hline
		45,000	&	43.2\%	&	884	&	98.8\%	& sniff, handout							\\\hline
		50,000	&	48.0\%	&	888	&	99.2\%	& unintellig, cheer, footnot, ridicul		\\\hline
		55,000	&	52.8\%	&	888	&	99.2\%	&											\\\hline
		60,000	&	57.6\%	&	888	&	99.2\%	&											\\\hline
		75,000	&	72.0\%	&	889	&	99.3\%	& nasti										\\\hline
		80,000	&	76.8\%	&	890	&	99.4\%	& parenthesi								\\\hline
		89,351	&	85.7\%	&	891	&	99.6\%	& whoever									\\\hline

		\end{tabular}
	\label{table:coverage2}
	\end{table}

An alternative view of fragment comparison is to evaluate the last position of the words of a specific fragment of the NAWL in LScDC.  Table \ref{table:coverage3} shows the fragment of NAWL and positions in LScDC. Both dictionaries are ranked by their frequencies: SFI for NAWL and the number of documents containing the word for the LScDC. We see that the first half of NAWL words is in approximately the first 14\% of LScDC. When the fragment of words in NAWL doubles, the position became around 5 times far from the first word in LScD.  We see that there are 300 words lies between 10,967th and 14,017th words in LScDC (100-400), with an interval of approximately 4,000 words. This interval is around 9,000 for the next 200 words, and followed by an interval of 55,000 for the third 200 words. Thus, we conclude that there are dense regions of LScDC in terms of the coverage of NAWL. 

	\begin{table}[h]
		\centering
		\caption{Last position of the words of a specific fragment of the NAWL in LScDC. Both dictionaries are sorted by their frequencies defined.}
		\renewcommand\arraystretch{1.3}
  			\begin{tabular}{| R{2.5cm} | R{4cm} | R{2.5cm} |}
	  	 	\hline
	 \multicolumn{1}{|C{2.5cm}|}{\textbf{Fragment of NAWL}} & \multicolumn{1}{C{4cm}|}{\textbf{Last position of  words of NAWL in LScDC}} & \multicolumn{1}{C{2.5cm}|}{\textbf{Fragment of LScD (\%)}} \\ \hline
 			
		100 	&	10,967 	&	10.5\%	 											\\\hline
		200 	&	10,967 	&	10.5\%												\\\hline
		300		&	10,967 	&	10.5\%												\\\hline
		400		&	14,017 	&	13.4\%											    \\\hline
		500		&	17,212 	&	16.5\%												\\\hline
		600		&	23,188 	&	22.2\%												\\\hline
		700		&	78,492 	&	75.3\%	 											\\\hline
		800		&	78,492 	&	75.3\%	 											\\\hline
		891		&	89,351 	&	85.7\%												\\\hline
		
		\end{tabular}
	\label{table:coverage3}
	\end{table}

\subsubsection{Comparison of Ranks of Words in Two Dictionaries \nopunct}\label{rank} \hspace*{\fill} \\

Our second approach to compare two lists is based on the order of words. The goal is to examine whether the ranking of words (frequency-based sorting) in dictionaries are actually similar. Note that only common words in both dictionaries (891 words) are taken into account. Words in both lists are descending ordered by their ranks in corpora, which are the number of documents containing the word in LScDC and SFI in NAWL. Table \ref{table:coverage4} shows stemmed versions of top 10 words with corresponding statistics in two lists. 

	\begin{table}[h]
		\centering
		\caption{The top 10 words in stemmed form with corresponding statistics in lists. Blue coloured words are matches in the top 10 of two dictionaries.}
		\renewcommand\arraystretch{1.3}
  			\begin{tabular}{| L{2.5cm} | R{2.5cm} | L{2.5cm} |  R{3cm}| }
	  	 	\hline
	 \multicolumn{1}{|C{2.5cm}|}{\textbf{Word in NAWL}} & \multicolumn{1}{C{2.5cm}|}{\textbf{SFI in NAWL}} & \multicolumn{1}{C{2.5cm}|}{\textbf{Word in LScDC}} & \multicolumn{1}{C{3cm}|}{\textbf{The number of documents containing the word in LScDC}} \\ \hline
 			
		repertoir 						&	72.45 	&	effect					& 		476,757 								\\\hline
		\textcolor{blue}{obtain}	 	&	66.52 	&	compar					&		355,381 								\\\hline
		distribut						&	65.67 	&	activ					&		255,630 								\\\hline
		paramet							&	64.37 	&	observ					&		249,965 							    \\\hline
		aspect							&	64.19 	&	found					&		234,720 								\\\hline
		dynam							&	63.51 	&	import					&		233,138 								\\\hline
		impact							&	63.49 	&	indic	 				&		229,775 								\\\hline
		domain							&	63.47 	&	demonstr				& 		218,861 								\\\hline
		publish							&	62.89 	&	\textcolor{blue}{obtain}&		218,578 								\\\hline
		denot							&	62.57 	&	condit					&		205,643 								\\\hline
		\end{tabular}
	\label{table:coverage4}
	\end{table}

From an inspection of order of words, 7 words in the lists is in the same order in both dictionaries when LScDC is restricted by NAWL words. Such words are listed in the Table \ref{table:coverage5}. Thus, the direct comparison of order cannot be used.

	\begin{table}[h]
		\centering
		\caption{Words in the same order in both dictionaries. The LScDC is restricted by NAWL words, we ignore other words to compare raking of words in dictionaries.}
		\renewcommand\arraystretch{1.3}
  			\begin{tabular}{| L{2cm} | R{2.5cm} | R{4cm} | R{2cm} |}
	  	 	\hline
	 \multicolumn{1}{|C{2cm}|}{\textbf{Word}} & \multicolumn{1}{C{2.5cm}|}{\textbf{Order of 
word in the lists}} & \multicolumn{1}{C{4cm}|}{\textbf{Number of documents containing the word in LScDC}} & \multicolumn{1}{C{2cm}|}{\textbf{SFI in NAWL}} \\ \hline
 			
		acut 		&	182 	&	30,876	& 	57.72 									\\\hline
		decay 		&	368 	&	12,761	& 	55.66									\\\hline
		horizon		&	543 	&	4,897	& 	53.83									\\\hline
		portfolio	&	656 	&	2,299	& 	52.13									\\\hline
		kilomet		&	778 	&	872		& 	49.14									\\\hline
		cheat		&	844 	&	310		& 	46.02									\\\hline
		handout		&	883 	&	51	 	& 	42.85									\\\hline
			
		\end{tabular}
	\label{table:coverage5}
	\end{table}

\begin{figure}[h]
\centering
 \includegraphics[width=0.65\linewidth]{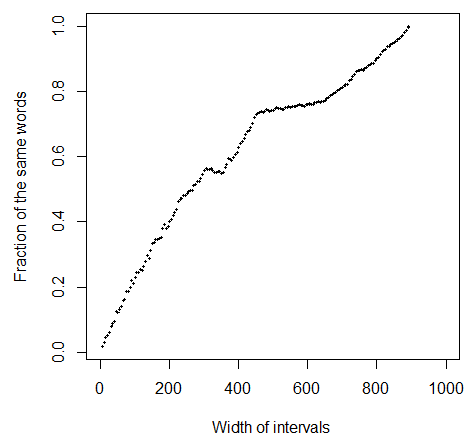}
   \caption{The fraction of the intersection of words in intervals with specified width.}
  \label{fig:widthfrac}
\end{figure}

\begin{figure}[h]
\centering
 \includegraphics[width=1\linewidth]{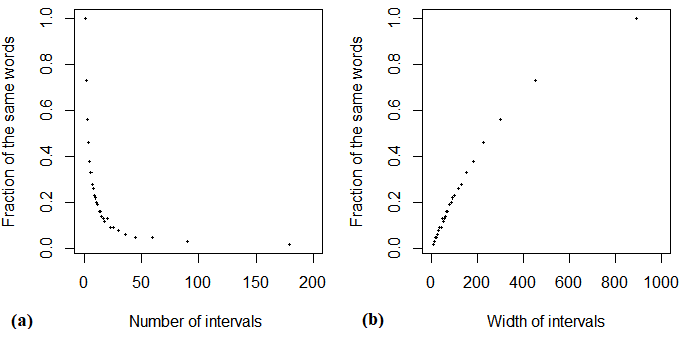}
   \caption{The fraction of the intersection of words in intervals with number of intervals and specified width. Figures present only those number of intervals and widths where partitions have almost equal widths (e.g. 2 intervals with approximately 450 words in each, 450 in one of intervals and 441 in the other interval).}
  \label{fig:intervalfrac}
\end{figure}

A new evaluation method is offered that focuses on pairwise comparison of partitions in dictionaries. The word lists are divided into smaller sub-lists, with the same number of intervals. We introduce an analysis that is focused on the overlapping words in intervals by counting the number of words in common. Within intervals, the common words are counted, and then the percentage of pairwise intersection of parts (total overlap) are considered to be the agreement of rating between LScDC and NAWL. As would expected, the larger width of intervals (small number of splits) yields the highest agreement of rating. The highest possible width is 891 (only 1 split) as there are 891 words in lists. To find the percentage of total overlap within intervals, the following statistical computation is done:

$$
\sum_{i}n_i \over N_t
$$
where $ n_i $ is the size of intersection in ith interval, and $ N_t $ is the total number of words (891). We repeated the same calculation for different widths of intervals, with an increasing sequence 5, 10, 15, … 890, 891. For instance, when the width is 5 the lists are divided into 179 intervals: 178 complete interval with 5 words, 1 shorter interval with 1 word. Figure \ref{fig:widthfrac} shows the fraction of the intersection in intervals with specified width. Observe that not in all cases lists are divided into equal intervals. For instance, the width 890 of interval means that there are two partitions with 890 and 1 words and so the comparison is not much meaningful in these cases. To avoid unbalanced classes, we consider only those number of intervals where partitions have almost equal widths. Figure \ref{fig:intervalfrac} and Table \ref{table:coverage6} show the number of intervals selected and the width of intervals for these intervals. When the lists are divided into two intervals, the fraction of overlap is 0.73. Hence, 27\% of words of a list do not lie within the same half of the other list. In addition, almost half of words are in different intervals when splitting the lists into 3 intervals, with approximately 300 words in each interval (300 words in two intervals and 291 words in one interval). Our findings raise the possibility that two lists are slightly different in terms of ranking words within lists. 

	\begin{table}[bt]
		\centering
		\caption{The percentage of the overlapping of words in intervals with number of intervals and width of intervals.}
		\renewcommand\arraystretch{1.3}
  			\begin{tabular}{| R{1.5cm} |  R{1.5cm} | R{2cm} |  R{1.5cm} |  R{1.5cm} | R{2cm} |}
	  	 	\hline
	 \multicolumn{1}{|C{1.5cm}|}{\textbf{Number of interval}} & \multicolumn{1}{C{1.5cm}|}{\textbf{Width of interval}} & \multicolumn{1}{C{2cm}|}{\textbf{Percentage of overlapping}} &  \multicolumn{1}{C{1.5cm}|}{\textbf{Number of interval}} & \multicolumn{1}{C{1.5cm}|}{\textbf{Width of interval}} & \multicolumn{1}{C{2cm}|}{\textbf{Percentage of overlapping}} \\ \hline
 			
179		&	5	&	1.8\%	&	13	&	70	&	16.2\%	\\ \hline
90		&	10	&	2.9\%	&	12	&	75	&	18.5\%	\\ \hline
60		&	15	&	4.5\%	&	11	&	85	&	20.0\%	\\ \hline
45		&	20	&	5.1\%	&	10	&	90	&	21.9\%	\\ \hline
36		&	257	&	6.2\%	&	9	&	100	&	22.9\%	\\ \hline
30		&	30	&	8.0\%	&	8	&	115	&	25.5\%	\\ \hline
26		&	35	&	8.6\%	&	7	&	130	&	27.9\%	\\ \hline
23		&	40	&	9.3\%	&	6	&	150	&	33.2\%	\\ \hline
20		&	45	&	12.6\%	&	5	&	180	&	37.9\%	\\ \hline
18		&	50	&	12.2\%	&	4	&	225	&	46.4\%	\\ \hline
17		&	55	&	13.2\%	&	3	&	300	&	55.7\%	\\ \hline
15		&	60	&	14.1\%	&	2	&	450	&	72.8\%	\\ \hline
14		&	65	&	15.8\%	&	1	&	891	&	100.0\%	\\ \hline

			\end{tabular}
	\label{table:coverage6}
	\end{table}

\subsubsection{Testing Similarity of Ranks in Two Dictionaries \nopunct}\label{testrank} \hspace*{\fill} \\

\begin{figure}[tb]
\centering
 \includegraphics[width=1\linewidth]{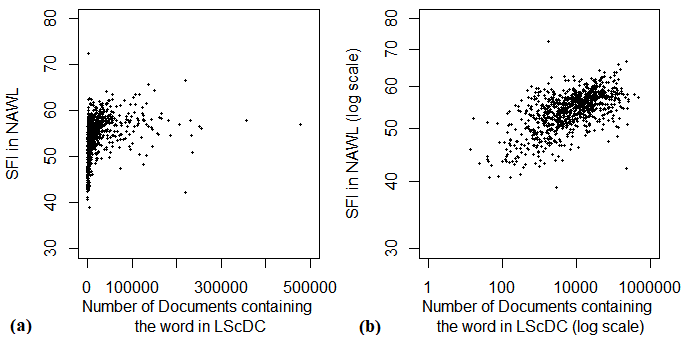}
   \caption{Relationship between the number of documents containing the word in LScD and SFIs in NAWL. The figure on the right hand side is on logarithmic scale.}
  \label{fig:sfi}
\end{figure}

The scatter plot suggests a positive correlation between frequencies in the LScDC and SFI values in the NAWL (see Figure \ref{fig:sfi}). In order to test whether there is any or no evidence to suggest that linear correlation of ranks is present in two dictionaries, the Spearman’s Rank Correlation (SRC) is used. Spearman’s correlation coefficient is a statistical measure of the strength and direction of a monotonic association between two ranked variables. It is actually equal to Pearson’s Correlation Coefficient (PCC) between two variables with ranked-values \cite{hauke}. 

\newpage
For a sample size $ n $, the Spearman’s coefficient $ R_s $ is computed as:

$$
R_s=\dfrac{1-6(\sum d_i^2)}{n^3-n}
$$
where $ d_i $ is the difference in the ranks of each variable pair \cite{dodge}.

\begin{figure}[h]
\centering
 \includegraphics[width=0.5\linewidth]{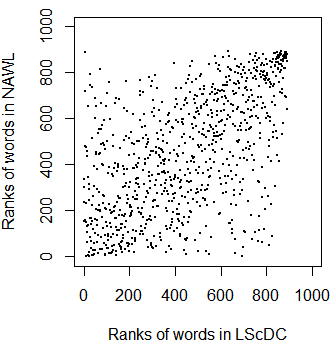}
   \caption{Relationship between ranks of words in lists (LScDC and NAWL).}
  \label{fig:rank}
\end{figure}

In this study, the Spearman’s correlation is calculated by assigning a rank of 1 to the highest value within each list, 2 to the next highest and so on. Figure \ref{fig:rank} presents the relationship between ranks of words in lists. The correlation between words in two lists will be high when words have a similar rank within lists. The calculation of Spearman correlation for this study gives a value of 0.58 which confirms what was found in the comparison of ranks and what was apparent from the graph. There is indeed a moderate positive correlation between two lists, which are monotonically related. We also calculated the Pearson’s correlation coefficient with frequencies and logarithmic scaled-frequencies, found 0.30 and 0.61 respectively (see Table \ref{table:corrolation}). This is expected results because we did not observe a linear relationship of frequencies, but monotonic in Figure \ref{fig:sfi}. However, the logarithmic scaled-frequencies show a linear relation.    

\begin{table}[h]
		\centering
		\caption{Correlation coefficients that measure the relationship between ranks of words in NAWL and LScDC: Pearson’s Correlation Coefficient (PCC), Spearman’s Rank Correlation (SRC) and PCC
 for logarithmic scaled frequencies.}
		\renewcommand\arraystretch{1.3}
  			\begin{tabular}{| L{2cm} |  R{3cm} | }
	  	 	\hline
	 \multicolumn{1}{|C{2cm}|}{\textbf{Test}} & \multicolumn{1}{C{3cm}|}{\textbf{Test Statistics}}  \\ \hline
 			
PCC			&	0.30			\\ \hline
SRC			&	0.58			\\ \hline
PCC-log		&	0.61			\\ \hline

			\end{tabular}
	\label{table:corrolation}
	\end{table}

\subsubsection{An Alternative Comparison of Ranks of Words in Two Dictionaries \nopunct}\label{rank2} \hspace*{\fill} \\

\begin{figure}[tb]
\centering
 \includegraphics[width=1\linewidth]{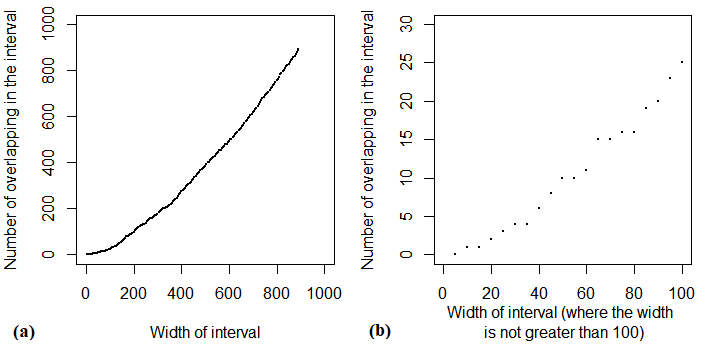}
   \caption{The number of overlapping of words in the top n words of the lists (width of interval) where $n=5,10,15,…,890,891$. Lists are in descending order by their statistics provided (the number of documents containing the word and SFI). The figure on right hand side presents widths until $n=100$.}
  \label{fig:overlapping1}
\end{figure}

Finally, we perform another analysis that is focused on ranks of words, similar to the comparison of ranks by partitioning intervals. Here, common words in both dictionaries (891 words) are used for analysis as in the previous comparison of ranks. The difference in this approach is the creation of intervals. Rather than dividing the whole lists into intervals, we consider the top n words by frequencies presented in dictionaries, where $n=5,10,15,…,890,891$ . For instance, if $n=5$ we compare the first 5 words in dictionaries, where words are ordered by the number of documents containing the word for LScDC and SFI for NAWL. Figure \ref{fig:overlapping1} shows the number of overlapping of words in top words for specified top n words. Note that in the figures, words are in descending order by their frequencies in both dictionary. We see that there are only 2 common words in the first frequent 20 words of lists. In the top 100 words, this number is 25, which means 25\% of words are common. This shows that the widely used words in corpora are slightly different. This may be result of the differences in calculations of statistics for words (the number of documents containing the word and SFI). 

\begin{figure}[tb]
\centering
 \includegraphics[width=1\linewidth]{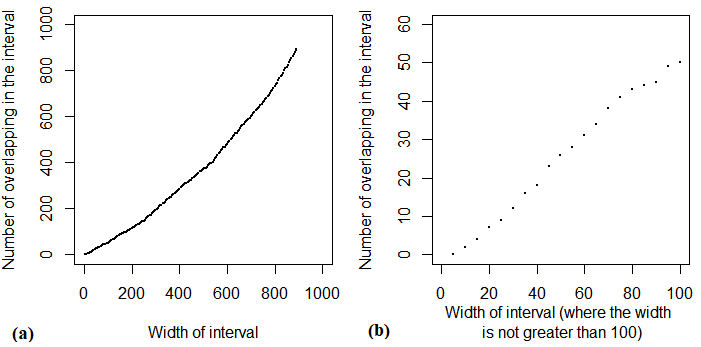}
   \caption{The number of overlapping of words in the bottom n words of the lists (width of interval) where $n=5,10,15,…,890,891$. Lists are in ascending order by their statistics provided (the number of documents containing the word and SFI). The figure on right hand side presents widths until $n=100$.}
  \label{fig:overlapping2}
\end{figure}

We repeated this analysis for ascending order of frequencies. In this case, we consider the bottom n words, where $n=5,10,15,…,890,891$. Figure \ref{fig:overlapping2} shows the number of overlapping of words for specified bottom n words. We can see that the number of overlapped words for bottom is much more when comparing top words. There are 7 common words in the least frequent 20 words of lists and 50 common words in the bottom 100 words (50\% of words). This means that there is an improvement in common words for the least frequent words. Dictionaries are more similar for bottom words. 

\section{Conclusion and Discussion} \label{conclusion}

In this work, we presented Leicester Scietific Corpus (LSC), Leicester Scientific Dictionary (LScD) and  Leicester Scientific Dictionary-Core (LScDC) with a description of the methodology and all steps in construction processes. Both the corpus and the dictionaries set out with the aim of quantifying the meaning in research texts in our future work.

LSC is a corpus of  abstracts of academic articles and proceeding papers, where all papers are indexed by WoS and published in 2014 in English. It consists of 1,673,824 abstracts with the metadata: title, list of authors, list of subject categories, list of research areas, times cited. In 119 documents, list of authors are not present; however, we did not exclude them. The average length of abstract is 178 words (all words including stop words and different forms of words) with a minimum 30 and a maximum 500 words. Each paper in WoS is assigned to at least one of subject categories and research areas. The number of subject categories that a paper is assigned to vary from 1 to 6 in the LSC.  

We then developed the LScD by extracting unique words (excluding stop words and various inflected forms of words) from the LSC. LScD is a scientific dictionary where all words are in stemmed form. It consists of 974,238 words; the number of documents containing each word and the number of appearance of each word in entire corpus are presented with the dictionary. Approximately 60\% of words appear in only one document, followed by 72\% for one or two documents. We observed that very few words occur very often, large number of mid-frequency words and very many words occur very rare (see Figure \ref{fig:docswords}). This indicates the Pareto's distribution behaviour. Pareto's law originally stated that `number of people with incomes higher than a certain limit follows a power law'. We can reword the law as `number of words appearing in more than a certain limit of document ($n$) follows a power law' (heavy-tailed distribution). The Pareto distribution is fitted the data with the Pareto index 0.5752 (see Figure \ref{fig:normalisedfrag}). 
 
LScDC is a core dictionary built by sub-setting the LScD. We decided to remove too rare words under the assumption that they do not contribute to the text categorisation and are likely to have noisy behaviour in the algorithms. Such words also have impact on measuring of meaning in texts by using the probabilistic approaches such as information gain. They are given almost zero score in such approaches. Therefore, we set a cut-off (10) to remove all words (in LScD) appearing in not greater than 10 documents in LSC. After removal of words, we obtained the LScDC containing 104,223 unique words. Words in LScDC, similar to the LScD, are associated with the number of documents containing the word and appearance of the word in entire corpus. 

Finally, we present a comprehensive analysis of LScDC by comparing with the NAWL. The NAWL is a list of academic words containing 973 word families. Our aim is to investigate how similar two lists are in terms of mainly matched words and the ranking of words. We applied many approaches based on both direct comparison of words and pairwise comparison of partitions of dictionaries in smaller subsets.  

Identification of the NAWL words in LScDC shows that out of the 895 word families (after applying stemming to the NAWL words) in NAWL, 891 were found to be included in LScDC, indicating that the LScDC represents almost complete NAWL words. Four words which appear in only NAWL are ``pi", ``ex", ``applaus" and ``pardon". These words did not appear in LScDC but in NAWL due to differences in pre-processing and types of texts in corpora.   

The ranking positions of many words of NAWL words found in LScDC are slightly different from those in the NAWL itself. We hypothesize that this is due to the difference in calculation of statistics used to order words in lists. In NAWL, the ordering of words is based on both the frequency and the dispersion of words over categories (SFI), while LScDC words are ordered by the number of documents containing the word only. From the plot of frequencies of matched words (SFI against the number of documents containing the word), we observed a monotonic relationship of frequencies (see Figure \ref{fig:sfi}). However, the log-log plot of matched words' statistics  suggests a positive correlation between statistics (linear relationship). We tested this similarity of rankings by Spearsman's Rank Correlation (SRC), Pearson's Correlation Coefficient (PCC) with statistics given and PCC with logarithmic scaled-statistics. We found correlation coefficients of 0.58, 0.30 and 0.61 respectively. This was indeed an expected result as it is the same what we observed from plots.  

We then perform an analysis on ranking positions of words by checking the overlap the top $n$ words in the LScDC and the NAWL successively where $n=5,10,15,...,890,891$.  We report that there are only 2 common words in the top 20 words of dictionaries, followed by 25 in the top 100 words. The same analysis was repeated for the bottom $n$ words and found that there are 7 common words in the least frequent 20 words, followed by 50 common words in the bottom 100 words. From these findings, we conclude that the LScDC and the NAWL are more similar for least frequent words. 
  
LSC is a multidisciplinary academic corpus of abstracts where the subject categories and citations are known. The dictionaries LScD and LScDC are scientific dictionaries where words are extracted from the LSC. This corpus and dictionaries will be used in a comprehensive research in quantification of meaning of research texts. The meaning of each word will be represented by an analysis of information on categories and areas of research that can be extracted from the appearance of this word in the text. Therefore, the next step will be measuring meaning in LSC texts and then using such measures in several data mining applications including prediction of success of the paper, categorisation of texts to pre-existing categories and clustering of texts into `natural categories'.

LSC, LScD and LScDC are available online in \cite{lsc,lscd,lscdc}.

\section{Acknowledgement}
We thank to Dr. Charles Browne, Dr. Brent Culligan and Joseph Phillips for their prompt replies to our emails and providing very useful information about the NAWL.

\section*{Appendices}
\appendix
\section{Table of Headings of Sections in Medical Abstracts} 
\raggedbottom
	\begin{table}[h]
		\centering
		\caption{Headings of sections identified in structured abstracts.}
		\renewcommand\arraystretch{1.5}
  			\begin{tabular}{ | l | l | l | l | l | l |  }
	  	 		\hline
\multicolumn{6}{|c|}{\textbf{Headings of Sections}} \\\hline
Abstract	& Aim(s)	 & Approach	      & Background    &  Conclusion(s)  & Design       \\ \hline
Discussion	& Finding(s) & Hypothesis     & Introduction  & Limitation(s)   & Location     \\ \hline
Material(s) & Measure(s) & Measurement(s) & Method(s)     & Methodology     & Objective(s) \\ \hline
Patient(s)	& Population & Procedure(s)   & Process       & Purpose(s)      & Rationale(s) \\ \hline
Result(s)   & Setting(s) & Subject(s)	  & Theoretical   &                 &              \\ \hline
\multicolumn{3}{|c|}{Implication(s) for health and nursing  policy} &       &           &  \\ \hline
			\end{tabular}
	 \label{table:sechead}
	\end{table}

\section{An Example of Document Structure in the LSC.}
\raggedbottom
\rotatebox{90}{
	\begin{minipage}{19cm}

\captionof{table}{Structure of a document in the LSC.} \label{docst}

\bigskip  

  			\begin{tabular}{ | L{3cm} | L{3cm} | L{4cm} | L{2cm} | L{2cm} | R{1cm} |  R{1cm} |}
	  	 		\hline
	  	 		\rc \textbf{Authors} & \rc \textbf{Title} & \rc \textbf{Abstract} & \rc \textbf{Categories}  &\rc \textbf{Research Areas} &\rc \textbf{Total Times Cited} &\rc \textbf{Times Cited in CC} \tn \hline
	  	 		
	  		   Cheng, JS; Craft, R; Yu, GQ; Ho, K; Wang, X; Mohan, G; Mangnitsky, S; Ponnusamy, R; Mucke, L &  Tau Reduction Diminishes Spatial Learning and Memory Deficits after Mild Repetitive Traumatic Brain Injury in Mice &  Objective: Because reduction of the microtubule-associated protein Tau has beneficial effects in mouse models of Alzheimer's disease and epilepsy, we wanted to determine whether this strategy can also improve the outcome of mild traumatic brain injury (TBI).  \textbf{ …(truncated)} &  Multidisciplinary Sciences & Science \& Technology - Other Topics & 24 &  24  \\ \hline
			
			\end{tabular}
	\end{minipage}}

\section{List of Categories.}
\raggedbottom	

\begin{longtable}[h]{| L{0.8cm} | L{8cm} |  R{2cm} | }
	
		\caption{The list of categories with the number of documents assigned to the corresponding category. There are 252 categories in the LSC. \label{table:categories}}\\ 
		\hline
  	  \multicolumn{1}{|C{0.8cm}|}{\textbf{No.}} & \multicolumn{1}{C{8cm}|}{\textbf{Category}} & \multicolumn{1}{C{2cm}|}{\textbf{Number of Documents}}\\ \hline
\endhead
\hline
\endfoot

1	&	Engineering, Electrical \& Electronic	   			&       174,305  \\ \hline
2	&	Materials Science, Multidisciplinary	   			&       112,920  \\ \hline
3	&	Physics, Applied	                       			&        78,824  \\ \hline
4	&	Chemistry, Physical	 					  			&        58,070  \\ \hline
5	&	Chemistry, Multidisciplinary	           			&        55,919  \\ \hline
6	&	Computer Science, Theory \& Methods	       			&        55,596  \\ \hline
7	&	Multidisciplinary Sciences	               			&        53,140  \\ \hline
8	&	Engineering, Mechanical	                   			&        50,972  \\ \hline
9	&	Optics	             					   			&  		 47,776  \\ \hline
10	&	Biochemistry \& Molecular Biology	       			&	     47,491  \\ \hline
11	&	Computer Science, Information Systems	   			&        45,867  \\ \hline
12	&	Energy \& Fuels	                           			&        44,202  \\ \hline
13	&	Environmental Sciences	                   			&        42,083  \\ \hline
14	&	Computer Science, Artificial Intelligence  			&        41,210  \\ \hline    
15	&	Telecommunications	                       			&        40,550  \\ \hline
16	&	Nanoscience \& Nanotechnology	           			&        35,052  \\ \hline
17	&	Oncology	      						  			&        34,340  \\ \hline
18	&	Mechanics	              				   			&		 33,550  \\ \hline
19	&	Neurosciences	             			   			&  		 32,974  \\ \hline
20	&	Surgery	                                   			& 		 30,818  \\ \hline
21	&	Pharmacology \& Pharmacy	               			& 		 30,714  \\ \hline
22	&	Automation \& Control Systems	           			&  		 29,429  \\ \hline
23	&	Engineering, Chemical	                    		&        29,172  \\ \hline
24	&	Computer Science, Interdisciplinary Applications	&     	 29,156  \\ \hline
25	&	Mathematics, Applied	             				& 		 28,105  \\ \hline
26	&	Physics, Condensed Matter	                        & 		 27,316  \\ \hline
27	&	Biotechnology \& Applied Microbiology	            & 		 26,286  \\ \hline
28	&	Mathematics	              							& 		 25,615  \\ \hline
29	&	Public, Environmental \& Occupational Health	    &        25,494  \\ \hline
30	&	Geosciences, Multidisciplinary	                    &		 24,644  \\ \hline
31	&	Cell Biology	             						&        23,108  \\ \hline
32	&	Physics, Multidisciplinary	                        &        22,932  \\ \hline
33	&	Astronomy \& Astrophysics	                        &        22,833  \\ \hline
34	&	Economics	                                        &		 22,343  \\ \hline
35	&	Clinical Neurology	             					&		 22,131	 \\ \hline
36	&	Engineering, Civil	             					&		 22,127	 \\ \hline
37	&	Chemistry, Analytical	        					&		 21,491  \\ \hline
38	&	Plant Sciences	             	 					&	 	 21,322  \\ \hline
39	&	Engineering, Multidisciplinary	 					&		 21,146  \\ \hline
40	&	Radiology, Nuclear Medicine \& Medical Imaging	    &        21,015  \\ \hline
41	&	Food Science \& Technology	                        &        20,414  \\ \hline
42	&	Education \& Educational Research	                &        20,088  \\ \hline
43	&	Medicine, Research \& Experimental	                &        19,744  \\ \hline
44	&	Genetics \& Heredity	                            &        19,512  \\ \hline
45	&	Computer Science, Hardware \& Architecture	        &        18,489  \\ \hline
46	&	Immunology	                                        &        18,270  \\ \hline
47	&	Chemistry, Organic	                                &        18,038  \\ \hline
48	&	Polymer Science	                                    &        18,017  \\ \hline
49	&	Engineering, Biomedical	              				&		 17,786  \\ \hline
50	&	Microbiology	                     				&		 17,252  \\ \hline
51	&	Computer Science, Software Engineering	            &        17,104  \\ \hline
52  &	Instruments \& Instrumentation	             		&        17,090  \\ \hline
53	&	Physics, Atomic, Molecular \& Chemical	            & 		 17,011  \\ \hline
54	&	Metallurgy \& Metallurgical Engineering	            &        16,899  \\ \hline
55	&	Ecology	             								&		 16,760  \\ \hline
56	&	Cardiac \& Cardiovascular Systems	             	&		 16,375  \\ \hline
57	&	Medicine, General \& Internal	             		&		 16,179  \\ \hline
58	&	Psychiatry	             							&		 16,056  \\ \hline
59	&	Electrochemistry	             					&		 15,664  \\ \hline
60	&	Biochemical Research Methods	             		&		 15,051  \\ \hline
61	&	Endocrinology \& Metabolism	              			&		 14,622  \\ \hline
62	&	Engineering, Environmental	             			&		 14,615  \\ \hline
63	&	Management	             							&		 14,339  \\ \hline
64	&	Chemistry, Applied	             					&		 14,060  \\ \hline
65	&	Water Resources	              						& 		 13,997  \\ \hline
66	&	Thermodynamics	             						&		 13,852  \\ \hline
67	&	Pediatrics	               							&		 13,370  \\ \hline
68	&	Physics, Particles \& Fields	             		&		 13,208  \\ \hline
69	&	Engineering, Manufacturing	             			&		 13,102  \\ \hline
70	&	Biophysics	             							&		 12,630  \\ \hline
71	&	Chemistry, Inorganic \& Nuclear	                 	&		 12,604  \\ \hline
72	&	Infectious Diseases	             					&		 12,524  \\ \hline
73	&	Chemistry, Medicinal	             				&		 12,463  \\ \hline
74	&	Meteorology \& Atmospheric Sciences	             	&		 12,319  \\ \hline
75	&	Construction \& Building Technology	             	&		 12,078  \\ \hline
76	&	Operations Research \& Management Science	        &     	 11,882  \\ \hline
77	&	Veterinary Sciences	             					&		 11,502  \\ \hline
78	&	Remote Sensing	             						&		 11,388  \\ \hline
79	&	Nuclear Science \& Technology	             		&		 11,360  \\ \hline
80	&	Zoology	             								&		 11,218  \\ \hline
81	&	Social Sciences, Interdisciplinary	             	&		 11,035  \\ \hline
82	&	Gastroenterology \& Hepatology	             		&		 10,943  \\ \hline
83	&	Orthopedics	             							&		 10,539  \\ \hline
84	&	Physics, Mathematical	             				&		 10,441  \\ \hline
85	&	Engineering, Industrial	             				&		 10,362  \\ \hline
86	&	Marine \& Freshwater Biology	             		&		 10,124  \\ \hline
87	&	Mathematics, Interdisciplinary Applications	        &    	 10,077  \\ \hline
88	&	Geochemistry \& Geophysics	             			&		 10,024  \\ \hline
89	&	Biology	                							&		  9,917  \\ \hline
90	&	Obstetrics \& Gynecology	                		&		  9,885  \\ \hline
91	&	Physics, Fluids \& Plasmas	                		&		  9,708  \\ \hline
92	&	Toxicology	                						&		  9,613  \\ \hline
93	&	Statistics \& Probability	                		&		  9,551  \\ \hline
94	&	Nutrition \& Dietetics	                     		&		  9,416  \\ \hline
95	&	Business	                						&		  9,394  \\ \hline
96	&	Imaging Science \& Photographic Technology	        &     	  9,354  \\ \hline
97	&	Hematology	               							&		  9,096  \\ \hline
98	&	Physiology	                						&		  9,009  \\ \hline
99	&	Peripheral Vascular Disease	                		&		  8,700  \\ \hline
100	&	Agronomy	                    					&		  8,651  \\ \hline
101	&	Dentistry, Oral Surgery \& Medicine	                &		  8,504  \\ \hline
102	&	Robotics	                						&		  8,491  \\ \hline
103	&	Transportation Science \& Technology	            &    	  8,412  \\ \hline
104	&	Sport Sciences	                					&		  8,368  \\ \hline
105	&	Psychology, Multidisciplinary	                	&		  8,333  \\ \hline
106	&	Urology \& Nephrology	                			&		  8,264  \\ \hline
107	&	Materials Science, Biomaterials	                	&		  8,040  \\ \hline
108	&	Mathematical \& Computational Biology	            &         8,015  \\ \hline
109 &	Health Care Sciences \& Services	                &         8,000  \\ \hline
110	&	Physics, Nuclear	                                &         7,886  \\ \hline
111	&	Ophthalmology	                                    &         7,832  \\ \hline
112	&	Environmental Studies	                            &         7,811  \\ \hline
113	&	Rehabilitation	                                    &         7,791  \\ \hline
114	&	Respiratory System	                                &         7,669  \\ \hline
115	&	Oceanography	                					&		  7,417  \\ \hline
116	&	Spectroscopy	                     				&         7,389  \\ \hline
117	&	Materials Science, Coatings \& Films	            &         7,226  \\ \hline
118	&	Pathology	                						&         7,217  \\ \hline
119	&	Business, Finance	                				&         7,214  \\ \hline
120	&	Psychology	                						&         6,989  \\ \hline
121	&	Acoustics	                						&         6,935  \\ \hline
122	&	Crystallography	               						&         6,935  \\ \hline
123	&	Psychology, Clinical	                            &		  6,860  \\ \hline
124	&	Geography, Physical	                                &		  6,806  \\ \hline
125	&	Psychology, Experimental	                		&		  6,784  \\ \hline
126	&	Nursing	                							&		  6,637  \\ \hline
127	&	Green \& Sustainable Science \& Technology	        &         6,412  \\ \hline
128	&	Agriculture, Multidisciplinary	                	&	      6,406  \\ \hline
129	&	Education, Scientific Disciplines	                &		  6,309  \\ \hline
130	&	Virology	                						&		  6,270  \\ \hline
131	&	Materials Science, Ceramics	               			&		  6,222  \\ \hline
132	&	Agriculture, Dairy \& Animal Science	            & 		  6,163  \\ \hline
133	&	Behavioral Sciences	                				&	      5,922  \\ \hline
134	&	Linguistics	                						&	      5,921  \\ \hline
135	&	Dermatology	               							&	      5,793  \\ \hline
136	&	Evolutionary Biology	                			&	      5,742  \\ \hline
137	&	Entomology	                						&	      5,705  \\ \hline
138	&	Parasitology	               						&		  5,683  \\ \hline
139	&	Horticulture	                					&		  5,338  \\ \hline
140	&	Health Policy \& Services	                		&		  5,318  \\ \hline
141	&	Language \& Linguistics	               				&		  5,174  \\ \hline
142	&	Political Science	                				&		  5,106  \\ \hline
143	&	Soil Science	               						&	      4,800  \\ \hline
144	&	Otorhinolaryngology	              					&	      4,797  \\ \hline
145	&	Geriatrics \& Gerontology	            			&	      4,743  \\ \hline
146	&	Sociology	                						&		  4,726  \\ \hline
147	&	Biodiversity Conservation	              			&		  4,705  \\ \hline
148	&	Fisheries	                						&		  4,702  \\ \hline
149	&	Engineering, Geological	                			&		  4,573  \\ \hline
150	&	Information Science \& Library Science	  			&         4,566  \\ \hline
151	&	Forestry	                						&		  4,472  \\ \hline
152	&	Engineering, Aerospace	               				&		  4,435  \\ \hline
153	&	Psychology, Developmental	               			&		  4,390  \\ \hline
154	&	Materials Science, Composites	               		&		  4,277  \\ \hline
155	&	Planning \& Development	               				&		  4,115  \\ \hline
156	&	Transplantation	                					&		  4,105  \\ \hline
157	&	Transportation	               						&		  4,036  \\ \hline
158	&	Medical Informatics	               					&		  3,992  \\ \hline
159	&	Reproductive Biology	                			&		  3,986  \\ \hline
160	&	Critical Care Medicine	               				&		  3,982  \\ \hline
161	&	Rheumatology	             						&	      3,942  \\ \hline
162	&	Geography	                						&		  3,908  \\ \hline
163	&	Materials Science, Characterization \& Testing	    &         3,878  \\ \hline
164	&	Agricultural Engineering	               			&		  3,727  \\ \hline
165	&	Tropical Medicine	              					&		  3,696  \\ \hline
166	&	Philosophy	               							&		  3,657  \\ \hline
167	&	Computer Science, Cybernetics	                	&	      3,652  \\ \hline
168	&	Developmental Biology	                			&	      3,594  \\ \hline
169	&	Law	               									&		  3,574  \\ \hline
170	&	Psychology, Social	              					&	      3,549  \\ \hline
171	&	Psychology, Applied	                				&		  3,523  \\ \hline 
172	&	Social Sciences, Mathematical Methods	            &         3,497  \\ \hline
173	&	History	               								&		  3,487  \\ \hline
174	&	Integrative \& Complementary Medicine	            &         3,453  \\ \hline
175	&	Substance Abuse	               						&	      3,433  \\ \hline
176	&	Communication	                					&		  3,200  \\ \hline
177	&	Anthropology	                					&		  3,150  \\ \hline
178	&	Social Sciences, Biomedical	              			&		  3,003  \\ \hline
179	&	Hospitality, Leisure, Sport \& Tourism	            &   	  2,998  \\ \hline
180	&	Anesthesiology	                					&		  2,943  \\ \hline
181	&	International Relations	                			&		  2,941  \\ \hline
182	&	Neuroimaging	               						&		  2,702  \\ \hline
183	&	Mining \& Mineral Processing	               		&		  2,687  \\ \hline
184	&	Emergency Medicine	                				&		  2,627  \\ \hline
185	&	Medical Laboratory Technology	               		&		  2,598  \\ \hline
186	&	Humanities, Multidisciplinary	                	&		  2,559  \\ \hline
187	&	Mineralogy	               							&	      2,550  \\ \hline
188	&	Materials Science, Textiles	                		&		  2,548  \\ \hline
189	&	Gerontology	                						&		  2,531  \\ \hline
190	&	Paleontology	               						&	      2,503  \\ \hline
191	&	Cell \& Tissue Engineering	                		&		  2,455  \\ \hline
192	&	Engineering, Ocean	                				&		  2,352  \\ \hline
193	&	Religion	               							&		  2,335  \\ \hline
194	&	Urban Studies	               						&		  2,309  \\ \hline
195	&	Family Studies	               						&		  2,229  \\ \hline
196	&	Public Administration	               				&		  2,204  \\ \hline
197	&	History \& Philosophy Of Science	               	&		  2,199  \\ \hline
198	&	Geology	                							&		  2,153  \\ \hline
199	&	Archaeology	               							&		  2,118  \\ \hline
200	&	Social Work	                						&		  2,114  \\ \hline
201	&	Psychology, Educational	                			&		  2,112  \\ \hline
202	&	Engineering, Marine	                				&		  2,110  \\ \hline
203	&	Audiology \& Speech-Language Pathology	            &         2,052  \\ \hline
204	&	Area Studies	                					&		  2,046  \\ \hline
205	&	Criminology \& Penology	                			&		  2,015  \\ \hline
206	&	Materials Science, Paper \& Wood	                &		  1,963  \\ \hline
207	&	Limnology	               							&		  1,941  \\ \hline
208	&	Engineering, Petroleum	               				&         1,930  \\ \hline
209	&	Ethics	                							&		  1,928  \\ \hline
210	&	Anatomy \& Morphology	                			&		  1,890  \\ \hline
211	&	Mycology	                						&		  1,829  \\ \hline
212	&	Logic	               								&	      1,791  \\ \hline
213	&	Allergy	                							&		  1,765  \\ \hline
214	&	Medicine, Legal	                					&		  1,712  \\ \hline
215	&	Education, Special	                				&		  1,666  \\ \hline
216	&	Literature	                						&		  1,608  \\ \hline
217	&	Psychology, Biological	                			&		  1,527  \\ \hline
218	&	Ergonomics	                						&		  1,431  \\ \hline
219	&	Architecture	                					&		  1,376  \\ \hline
220	&	Women's Studies	                					&		  1,341  \\ \hline
221	&	Microscopy	                						&		  1,319  \\ \hline
222	&	Social Issues	                					&		  1,296  \\ \hline
223	&	Primary Health Care	                				&		  1,269  \\ \hline
224	&	Ornithology	                						&		  1,008  \\ \hline
225	&	Cultural Studies	                   				&		    948  \\ \hline
226	&	Demography	                   						&		    948  \\ \hline
227	&	Music	                   							&			888  \\ \hline
228	&	Agricultural Economics \& Policy	                &  			880  \\ \hline
229	&	History Of Social Sciences	                   		&			879  \\ \hline
230	&	Industrial Relations \& Labor	                   	&			879  \\ \hline
231	&	Asian Studies	                   					&			877  \\ \hline
232	&	Art	                   								&			725  \\ \hline
233	&	Ethnic Studies	                   					&			675  \\ \hline
234	&	Medical Ethics	                   					&			674  \\ \hline
235	&	Psychology, Mathematical	                   		&			538  \\ \hline
236	&	Literary Theory \& Criticism	                   	&			498  \\ \hline
237	&	Medieval \& Renaissance Studies	                  	&			485  \\ \hline
238	&	Film, Radio, Television	                   			&			398  \\ \hline
239	&	Andrology	                   						&			391  \\ \hline
240	&	Psychology, Psychoanalysis	                   		&			345  \\ \hline
241	&	Classics	                   						&			325  \\ \hline
242	&	Theater	                   							&			300  \\ \hline
243	&	Literature, Romance	                   				&			269  \\ \hline
244	&	Literature, British Isles	                   		&			220  \\ \hline
245	&	Folklore	                   						&			134  \\ \hline
246	&	Literature, German, Dutch, Scandinavian	            &     		128  \\ \hline
247 &	Literature, American	                      		&			 75  \\ \hline
248 &	Dance	                      						&			 74  \\ \hline
249 &	Literature, African, Australian, Canadian	        &            59  \\ \hline
250 &	Poetry	                      						&			 42  \\ \hline
251 &	Literary Reviews	                      			&			 35  \\ \hline
252 &	Literature, Slavic	                                &            35  \\ \hline
 
	 \end{longtable}

\section{List of Research Areas.}
\raggedbottom	

\begin{longtable}[h]{| L{1cm}| L{8cm} |  R{2cm} | }
	
		\caption{The list of research areas with the number of documents assigned to the corresponding research area. There are 151 research areas in the LSC. \label{table:researcharea}}\\ 
		\hline
  	  \multicolumn{1}{|C{1cm}|}{\textbf{No.}} & \multicolumn{1}{C{8cm}|}{\textbf{Research Area}} & \multicolumn{1}{C{2cm}|}{\textbf{Number of Documents}}\\ \hline
\endhead
\hline
\endfoot

1	&	Engineering	   										&       328,173   \\ \hline
2	&	Chemistry	   										&       163,052   \\ \hline
3	&	Physics        										&       158,496   \\ \hline
4	&	Computer Science									&       142,642   \\ \hline
5	&	Materials Science	               					&       141,762   \\ \hline
6	&	Science \& Technology - Other Topics	            &        96,395   \\ \hline
7	&	Environmental Sciences \& Ecology	             	&  		 60,658   \\ \hline
8	&	Biochemistry \& Molecular Biology	       			&	     60,029   \\ \hline
9	&	Mathematics	   										&        59,757   \\ \hline
10	&	Neurosciences \& Neurology	                        &        48,684   \\ \hline
11	&	Optics	                   							&        47,776   \\ \hline
12	&	Energy \& Fuels  									&        44,202   \\ \hline    
13	&	Business \& Economics	                       		&        40,748   \\ \hline
14	&	Telecommunications	           						&        40,550   \\ \hline
15	&	Pharmacology \& Pharmacy	      					&        38,844   \\ \hline
16	&	Psychology	              				   			&		 36,284   \\ \hline
17	&	Oncology	             			   				&  		 34,340   \\ \hline
18	&	Mechanics	                                   		& 		 33,550   \\ \hline
19	&	Agriculture	               							& 		 31,191   \\ \hline
20	&	Surgery												& 		 30,818   \\ \hline
21	&	Automation \& Control Systems	           			&  		 29,429   \\ \hline
22	&	Geology	                    						&        26,632   \\ \hline
23	&	Biotechnology \& Applied Microbiology				&     	 26,286   \\ \hline
24	&	Education \& Educational Research	             	& 		 25,926   \\ \hline
25	&	Public, Environmental \& Occupational Health	    &        25,494   \\ \hline
26	&	Cell Biology	             						&        24,145   \\ \hline
27	&	Cardiovascular System \& Cardiology	                &        23,402   \\ \hline
28	&	Astronomy \& Astrophysics	                        &        22,833   \\ \hline
29	&	Plant Sciences	                                    &		 21,322   \\ \hline
30	&	Radiology, Nuclear Medicine \& Medical Imaging	    &		 21,015   \\ \hline
31	&	Food Science \& Technology	             			&		 20,414   \\ \hline
32	&	General \& Internal Medicine	        			&		 20,409   \\ \hline
33	&	Research \& Experimental Medicine	             	&	 	 19,744   \\ \hline
34	&	Genetics \& Heredity	 							&		 19,512   \\ \hline
35	&	Immunology	    									&        18,270   \\ \hline
36	&	Polymer Science	                        			&        18,017   \\ \hline
37	&	Microbiology	                					&        17,252   \\ \hline
38	&	Instruments \& Instrumentation	               		&        17,090   \\ \hline
39	&	Metallurgy \& Metallurgical Engineering	            &        16,899   \\ \hline
40	&	Social Sciences - Other Topics	       				&        16,666   \\ \hline
41	&	Psychiatry	                                        &        16,056   \\ \hline
42	&	Electrochemistry	                                &        15,664   \\ \hline
43	&	Endocrinology \& Metabolism	                        &        15,013   \\ \hline
44	&	Water Resources	              						&		 13,997   \\ \hline
45	&	Thermodynamics	                     				&		 13,852   \\ \hline
46	&	Pediatrics	            							&        13,370   \\ \hline
47	&	Biophysics	             							&        12,630   \\ \hline
48	&	Infectious Diseases	            					& 		 12,524   \\ \hline
49	&	Meteorology \& Atmospheric Sciences	            	&        12,319   \\ \hline
50	&	Zoology	             								&		 12,200   \\ \hline
51	&	Construction \& Building Technology	             	&		 12,078   \\ \hline
52	&	Operations Research \& Management Science	        &		 11,882   \\ \hline
53	&	Marine \& Freshwater Biology	             		&		 11,562   \\ \hline
54	&	Veterinary Sciences	             					&		 11,502   \\ \hline
55	&	Remote Sensing	             						&		 11,388   \\ \hline
56	&	Nuclear Science \& Technology	              		&		 11,360   \\ \hline
57	&	Gastroenterology \& Hepatology	             		&		 10,943   \\ \hline
58	&	Orthopedics	             							&		 10,539   \\ \hline
59	&	Transportation	             						&		 10,281   \\ \hline
60	&	Health Care Sciences \& Services              		& 		 10,244   \\ \hline
61	&	Geochemistry \& Geophysics	             			&		 10,024   \\ \hline
62	&	Life Sciences \& Biomedicine - Other Topics	        &		  9,917   \\ \hline
63	&	Obstetrics \& Gynecology	             			&		  9,885   \\ \hline
64	&	Toxicology	             							&		  9,613   \\ \hline
65	&	Nutrition \& Dietetics	             				&		  9,416   \\ \hline
66	&	Imaging Science \& Photographic Technology	        &		  9,354   \\ \hline
67	&	Hematology	             							&		  9,096   \\ \hline
68	&	Physiology	             							&		  9,009   \\ \hline
69	&	Dentistry, Oral Surgery \& Medicine	             	&		  8,504   \\ \hline
70	&	Government \& Law             						&		  8,492   \\ \hline
71	&	Robotics	        								&     	  8,491   \\ \hline
72	&	Sport Sciences	             						&		  8,368   \\ \hline
73	&	Urology \& Nephrology	             				&		  8,264   \\ \hline
74	&	Mathematical \& Computational Biology	            &		  8,015   \\ \hline
75	&	Ophthalmology	             						&		  7,832   \\ \hline
76	&	Rehabilitation	             						&		  7,791   \\ \hline
77	&	Respiratory System	             					&		  7,669   \\ \hline
78	&	Oceanography	             						&		  7,417   \\ \hline
79	&	Spectroscopy	             						&		  7,389   \\ \hline
80	&	Pathology	             							&		  7,217   \\ \hline
81	&	Linguistics	             							&		  7,077   \\ \hline
82	&	Acoustics	        								&    	  6,935   \\ \hline
83	&	Crystallography	             						&		  6,935   \\ \hline
84	&	Physical Geography	                				&		  6,806   \\ \hline
85	&	Nursing	                							&		  6,637   \\ \hline
86	&	Virology	                						&		  6,270   \\ \hline
87	&	Public Administration	                			&		  6,120   \\ \hline
88	&	Behavioral Sciences	                				&		  5,922   \\ \hline
89	&	Dermatology	               							&	      5,793   \\ \hline
90	&	Evolutionary Biology	                			&	      5,742   \\ \hline
91	&	Entomology	                						&	      5,705   \\ \hline
92	&	Parasitology	               						&		  5,683   \\ \hline
93	&	Geriatrics \& Gerontology	               			&	      5,506   \\ \hline
94	&	Otorhinolaryngology	              					&	      4,797   \\ \hline
95	&	Sociology	                						&		  4,726   \\ \hline
96	&	Biodiversity \& Conservation	              		&		  4,705   \\ \hline
97	&	Fisheries	                						&		  4,702   \\ \hline
98	&	Information Science \& Library Science	  			&         4,566   \\ \hline
99	&	Forestry	                						&		  4,472   \\ \hline
100	&	Transplantation	                					&		  4,105   \\ \hline
101	&	Medical Informatics	               					&		  3,992   \\ \hline
102	&	Reproductive Biology	                			&		  3,986   \\ \hline
103	&	Rheumatology	             						&	      3,942   \\ \hline
104	&	Geography	                						&		  3,908   \\ \hline
105	&	Tropical Medicine	              					&		  3,696   \\ \hline
106	&	Philosophy	               							&		  3,657   \\ \hline
107	&	Developmental Biology	                			&	      3,594   \\ \hline
108	&	Mathematical Methods In Social Sciences	            &         3,497   \\ \hline
109	&	History	               								&		  3,487   \\ \hline
110	&	Integrative \& Complementary Medicine	            &         3,453   \\ \hline
111	&	Substance Abuse	               						&	      3,433   \\ \hline
112	&	Communication	                					&		  3,200   \\ \hline
113	&	Arts \& Humanities - Other Topics					&		  3,178   \\ \hline
114	&	Anthropology	                					&		  3,150   \\ \hline
115	&	Biomedical Social Sciences	              			&		  3,003   \\ \hline
116	&	Anesthesiology	                					&		  2,943   \\ \hline
117	&	International Relations	                			&		  2,941   \\ \hline
118	&	Literature		               						&		  2,735   \\ \hline
119	&	Mining \& Mineral Processing	               		&		  2,687   \\ \hline
120	&	Emergency Medicine	                				&		  2,627   \\ \hline
121	&	Medical Laboratory Technology	               		&		  2,598   \\ \hline
122	&	Mineralogy	               							&	      2,550   \\ \hline
123	&	Paleontology	               						&	      2,503   \\ \hline
124	&	Religion	               							&		  2,335   \\ \hline
125	&	Urban Studies	               						&		  2,309   \\ \hline
126	&	Family Studies	               						&		  2,229   \\ \hline
127	&	History \& Philosophy Of Science	               	&		  2,199   \\ \hline
128	&	Archaeology	               							&		  2,118   \\ \hline
129	&	Social Work	                						&		  2,114   \\ \hline
130	&	Audiology \& Speech-Language Pathology	            &         2,052   \\ \hline
131	&	Area Studies	                					&		  2,046   \\ \hline
132	&	Criminology \& Penology	                			&		  2,015   \\ \hline
133	&	Anatomy \& Morphology	                			&		  1,890   \\ \hline
134	&	Mycology	                						&		  1,829   \\ \hline
135	&	Allergy	                							&		  1,765   \\ \hline
136	&	Legal Medicine	                					&		  1,712   \\ \hline
137	&	Architecture	                					&		  1,376   \\ \hline
138	&	Women's Studies	                					&		  1,341   \\ \hline
139	&	Microscopy	                						&		  1,319   \\ \hline
140	&	Social Issues	                					&		  1,296   \\ \hline
141	&	Cultural Studies	                   				&		    948   \\ \hline
142	&	Demography	                   						&		    948   \\ \hline
143	&	Music	                   							&			888   \\ \hline
144	&	Asian Studies	                   					&			877   \\ \hline
145	&	Art	                   								&			725   \\ \hline
146	&	Ethnic Studies	                   					&			675   \\ \hline
147	&	Medical Ethics	                   					&			674   \\ \hline
148	&	Film, Radio, Television	                   			&			398   \\ \hline
149	&	Classics	                   						&			325   \\ \hline
150	&	Theater	                   							&			300   \\ \hline
151	&	Dance	                      						&			 74   \\ \hline
	
		\end{longtable}	
	
\section{Lists of Prefixes and Substitutes.}
\raggedbottom

\begin{table}[h]
		\centering
		\caption{The List of Prefixes.}
		\renewcommand\arraystretch{1.5}
  			\begin{tabular}{ | l | l | l | l | l | l | l | }
	  	 		\hline
				\multicolumn{7}{|c|}{\textbf{Prefixes}} \\\hline
			anti- 	&  ante-   &  auto-	  &  co-    &  de-     &  deca-   &  di-     \\ \hline
			dia-   	&  dis-    &  e-      &  ex-    &  extra-  &  fore-   &  hemi-   \\ \hline
		    hexa-  	&  hepta-  &  homo-	  &  hyper- &  in-     &  inter-  &  im-     \\ \hline
		    ir-  	&  kilo-   &  micro-  &  mid-   &  milli-  &  mis-    &  mono-   \\ \hline
			multi-	&  non-    &  octo-   &  over-	&  para-   &  penta-  &  per-    \\ \hline
			poly-	&  post-   &  pre-    &  pro-   &  quadri- &  re-     &  retro-  \\ \hline
			self-	&  semi-   &  sub-    &  super- &  tele-   &  tetra-  &  therm-  \\ \hline
			trans-  & tri-     &  ultra-  & un-     & under-   &  uni-    &          \\ \hline
			\end{tabular}
	\label{table:prefix}
	\end{table}

\begin{table}[h]
		\centering
		\caption{The List of Substitution.}
		\renewcommand\arraystretch{1.5}
  			\begin{tabular}{ | l | l | }
	  	 		\hline
				    \textbf{Word} 	   & 	\textbf{Substitute}   \\ \hline
					well-known & 	wellknown	 \\ \hline
					z-test 	   &    ztest        \\ \hline
				    z-testing  &    ztest        \\ \hline
					z-tests    &    ztest        \\ \hline
					z-score    &    zscore	     \\ \hline
					z-scored   &    zscored      \\ \hline
					z-scores   &    zscore 	     \\ \hline
					p-value    &    pvalue       \\ \hline
					p-values   &    pvalue       \\ \hline
					p-valued   &    pvalue       \\ \hline
					p-valuesof &    pvalue       \\ \hline
					chi-square &    chisquare    \\ \hline
					chi-squares&    chisquare    \\ \hline
					chi-squared&    chisquared   \\ \hline
					chi2-test  &    chisquared   \\ \hline

			\end{tabular}
	\label{table:subs}
	\end{table}

\section{List of stop words in “tm” package (R package).}
\raggedbottom
\begin{table}[h]
		\centering
		\caption{The List of Stop Words.}
		\renewcommand\arraystretch{1.5}
	
  			\begin{tabular}{ | l | l | l | l | l | l | l | l | }
	  	 	\hline
	  	 	\multicolumn{8}{|c|}{\textbf{Stop Words in ‘tm’ Package}} \\ \hline
	  	i	     &  me	      &   my	      & myself	  &  we	       & our    &	ours	  &  ourselves      \\ \hline
		yours	 &  yourself  &	  yourselves  &	he        &  him	   & his    &	himself	  &  she	        \\ \hline
		herself	 &  it	      &   its	      & itself    &  they	   & them   &	their	  &  theirs	        \\ \hline
		which    &	who	      &   whom	      & this	  &  that	   & these	&   those	  &  am     	    \\ \hline
		was	     &  were	  &   be	      & been      &  being     & have   &	has       &	 had	        \\ \hline
		does	 &  did	      &   doing	      & would	  &  should    & could	&   ought	  &  i'm	        \\ \hline
		she's	 &  it's	  &   we're	      & they're	  &  i've	   & you've	&   we've	  &  they've        \\ \hline
		he'd	 &  she'd	  &   we'd	      & they'd	  &  i'll	   & you'll	&   he'll	  &  she'll	        \\ \hline
		isn't	 &  aren't	  &   wasn't	  & weren't	  &  hasn't    & haven't&	hadn't	  &  doesn't	    \\ \hline
		won't	 &  wouldn't  &	  shan't	  & shouldn't &  can't	   & cannot &	couldn't  &	 mustn't        \\ \hline
		who's	 &  what's	  &   here's	  & there's   &  when's	   & where's&	why's	  &  how's	        \\ \hline
        the      &	and	      &   but	      & if	      &  or        & because&   as        &	 until          \\ \hline
        at       &	by	      &   for	      & with	  &  about	   & against&	between   &	 into	        \\ \hline
        before	 &  after	  &   above	      & below	  &  to        & from	&   up	      &  down           \\ \hline
        on       &	off	      &   over        & under     &	 again	   & further&   then      &  once           \\ \hline
        when	 &  where     &	  why	      & how       &	 all       & any    &	both      &	 each           \\ \hline
        most	 &  other	  &   some        &	such      &	 no        & nor	&   not       &	 only           \\ \hline
        you      &	your      &   her	      & hers      &  themselves& what   &   is        &	 are            \\ \hline
        having	 &  do        &   you're	  & he's      &	 i'd       & you'd  &   we'll	  &  they'll        \\ \hline
        don't	 &  didn't    &	  let's	      & that's    &  a         & an     &	while	  &  of  			\\ \hline
        through  &	during    &	  in          &	out       &	 here	   & there  &	few       &	 more 			\\ \hline
        so	     &  than	  &   too	      & very	  &  own       & same   &             &	                \\ \hline			
	  	 
			\end{tabular}
	\label{table:stopwords}
	\end{table}

\end{document}